\documentclass[]{interact}

\usepackage{epstopdf}

\usepackage{natbib}
\bibpunct[, ]{(}{)}{,}{a}{}{,}
\renewcommand\bibfont{\fontsize{10}{12}\selectfont}

\usepackage{xcolor}

\usepackage[utf8]{inputenc} 
\usepackage[T1]{fontenc}    
\usepackage{hyperref}       
\usepackage{url}            
\usepackage{booktabs}       
\usepackage{amsfonts}       
\usepackage{nicefrac}       
\usepackage{microtype}      
\usepackage{lineno}
\usepackage{subcaption}
\usepackage{multirow}
\usepackage{graphicx}

\usepackage{makecell}

\theoremstyle{plain}
\newtheorem{theorem}{Theorem}[section]

\theoremstyle{definition}
\newtheorem{definition}[theorem]{Definition}

\newtheorem{property}{Property}[section]

\theoremstyle{remark}

\usepackage{changes}
\usepackage{lipsum}
\usepackage{adjustbox}

\usepackage{amsthm}
\usepackage{amsmath}
\usepackage{mathabx}
\usepackage{xspace}
\usepackage{color}
\usepackage{pgfplots, pgfplotstable}

\renewcommand{\vec}[1]{\boldsymbol{\mathbf{#1}}}

\def\bi{\vec{i}}

\def\bv{\vec{v}}

\def\bx{\vec{x}}

\def\mP{\mathcal{P}}
\def\mQ{\mathcal{Q}}
\def\Real{\mathbb{R}}

\newcommand{\spdim}{L}
\newcommand{\pt}{p}
\newcommand{\fedim}{E}

\newcommand{\embdim}{d}

\newcommand{\enc}{Enc}

\newcommand{\psenc}{PE}
\newcommand{\psdim}{W}

\newcommand{\pemlp}{\mathbf{NN}}

\def\lon{\lambda}
\def\lat{\phi}

\newcommand{\discretize}{discretize}
\newcommand{\onehot}{onehot}
\newcommand{\tile}{tile}

\newcommand{\kernel}{kernel}
\newcommand{\rbf}{rbf}
\newcommand{\kernelfunc}{k}

\newcommand{\numkernel}{q}

\newcommand{\kerneldecomp}{\Psi}

\newcommand{\gpsvec}{GPS2Vec}
\newcommand{\regularkernel}{\zeta}
\newcommand{\adpativekernelwidth}{h_{\sigma}}
\newcommand{\adpativekernel}{ak}

\newcommand{\direct}{direct}

\newcommand{\sinusoidal}{sinu}
\newcommand{\aodha}{wrap}

\newcommand{\sinmulti}{sinmul}
\newcommand{\grid}{grid}
\newcommand{\theory}{theory}

\newcommand{\nscale}{S}
\newcommand{\certainscale}{s}

\newcommand{\maxscale}{\lambda_{max}}
\newcommand{\minscale}{\lambda_{min}}
\newcommand{\scaleratio}{g}
\newcommand{\unit}{\mathbf{a}}
\newcommand{\theofuc}{\mathbf{\Psi}}

\newcommand{\protein}{TF}

\newcommand{\aggfunc}{Agg}
\newcommand{\comfunc}{Cmb}
\newcommand{\readoutfunc}{Rdt}

\newcommand{\numagglayer}{M}
\newcommand{\kthagglayer}{m}

\newcommand{\locemb}{\vec{h}}

\newcommand{\globalagg}{global}

\newcommand{\aggemb}{\mathbf{g}}
\newcommand{\poinet}{pnet}
\newcommand{\transnet}{TN}
\newcommand{\fcn}{FCN}

\newcommand{\singleagg}{local}
\newcommand{\buffer}{buffer}
\newcommand{\buffradius}{r}

\newcommand{\voxnet}{vox}
\newcommand{\voxlocate}{\iota}

\newcommand{\dgcnn}{dgcnn}
\newcommand{\dgcnnmodel}{DGCNN}
\newcommand{\edgeconv}{EdgeConv}
\newcommand{\locembset}{\mathcal{H}}

\newcommand{\sagatmodel}{SAGAT}
\newcommand{\sagat}{sagat}

\newcommand{\attnscr}{\alpha}
\newcommand{\attnvec}{\mathbf{a}}
\newcommand{\attnsize}{U}
\newcommand{\attnidx}{u}
\newcommand{\attnact}{\sigma} 

\newcommand{\leakyrelu}{LeakyReLU}

\newcommand{\knn}{knn}
\newcommand{\KNNfunc}{KNN}
\newcommand{\topk}{K}

\newcommand{\gcgan}{gcgan}

\newcommand{\multiagg}{hieagg}
\newcommand{\ptconv}{PTConv}
\newcommand{\ptdeconv}{PTDeConv}

\newcommand{\poinetplus}{pnet+}
\newcommand{\ssg}{ssg}

\newcommand{\maxpool}{MaxPool}
\newcommand{\mlp}{MLP}

\newcommand{\neifunc}{\mathcal{N}}

\newcommand{\pointcnn}{ptcnn}

\newcommand{\xtrans}{\mathcal{X}}

\newcommand{\rowconcat}{\Gamma}

\newcommand{\conv}{Conv}
\newcommand{\convkernel}{\kappa}

\newcommand{\dilate}{\mathcal{D}}

\newcommand{\graph}{\mathcal{G}}
\newcommand{\nodeset}{\mathcal{V}}
\newcommand{\edgeset}{\mathcal{E}}

\newcommand{\node}{e}

\newcommand{\mai}[1]{\textcolor{black}{{#1}}}

\begin{document}

\articletype{Review Paper}


\title{A Review of Location Encoding for GeoAI: Methods and Applications} 

\author{
\name{Gengchen Mai\textsuperscript{a,b,c}\thanks{CONTACT Gengchen Mai. Email: gengchen\_mai@geog.ucsb.edu}, Krzysztof Janowicz\textsuperscript{a,b}, Yingjie Hu\textsuperscript{d}, Song Gao\textsuperscript{e}, Bo Yan\textsuperscript{a,b}, Rui Zhu\textsuperscript{a,b}, Ling Cai\textsuperscript{a,b}, and Ni Lao\textsuperscript{f}
}
\affil{
\textsuperscript{a}STKO Lab, Department of Geography, UC Santa Barbara, CA, USA, 93106; \\
\textsuperscript{b}Center for Spatial Studies, UC Santa Barbara, CA, USA, 93106; \\
\textsuperscript{c}Department of Computer Science, Stanford University, CA, USA, 94305; \\
\textsuperscript{d}GeoAI Lab, Department of Geography, University at Buffalo, NY, USA, 14260; \\
\textsuperscript{e}GeoDS Lab, Department of Geography, University of Wisconsin-Madison, WI, USA, 53706; \\
\textsuperscript{f}Mosaix.ai, Palo Alto, CA, USA, 94303
}
}

\maketitle

\begin{abstract}
A common need for artificial intelligence models in the broader geoscience
is to represent and encode various types of spatial data, such as points (e.g., points of interest), polylines (e.g., trajectories), polygons (e.g., administrative regions), graphs (e.g., transportation networks), or rasters (e.g., remote sensing images), in a hidden embedding space so that they can be readily incorporated into deep learning models. One fundamental step is to encode a single point location into an embedding space, such that this embedding is learning-friendly for downstream machine learning models such as support vector machines and neural networks. We call this process \textit{location encoding}. However, there lacks a systematic review on the concept of location encoding, its potential applications, and key challenges that need to be addressed. This paper aims to fill this gap. We first provide a formal definition of location encoding, and discuss the necessity of location encoding for GeoAI research from a machine learning perspective. Next, we provide a comprehensive survey and discussion about the current landscape of location encoding research. We classify location encoding models into different categories based on their inputs and encoding methods, and compare them based on whether they are parametric, multi-scale, distance preserving, and direction aware. We demonstrate that existing location encoding models can be \textit{unified} under a shared formulation framework. We also discuss the application of location encoding for different types of spatial data. Finally, we point out several challenges in location encoding research that need to be solved in the future. 
\end{abstract}



\section{Introduction and Motivation}   \label{sec:intro}

The rapid development of novel deep learning and representation learning techniques and the increasing availability of diverse, large-scale geospatial data have fueled substantial progress in geospatial artificial intelligence (GeoAI) research \citep{smith1984artificial,couclelis1986artificial,openshaw1997artificial,janowicz2020geoai}. This includes progress on a wide spectrum of challenging tasks such as terrain feature detection and extraction \citep{li2020automated}, land use classification \citep{zhong2019deep},  navigation in the urban environment \citep{mirowski2018learning}, image geolocalization \citep{weyand2016planet,izbicki2019exploiting}, toponym recognition and disambiguation \citep{delozier2015gazetteer,wang2020neurotpr}, geographic knowledge graph completion and summarization \citep{qiu2019knowledge,yan2019spatially}, traffic forecasting \citep{li2017diffusion}, to name a few.

Despite the fact that these models are very different in design, they share a common characteristic - 
they need to \textit{represent} (or \textit{encode}) different types of spatial data, such as points (e.g., points of interest (POIs)), polylines (e.g., trajectories), polygons (e.g., administrative regions), graphs/networks (e.g., transportation networks), or raster (e.g., satellite images), in a hidden embedding space so that they can be utilized by machine learning models such as deep neural nets (NN). For raster data, this encoding process is straightforward since the regular grid structures can be directly handled by existing deep learning models such as convolutional neural networks (CNN) \citep{krizhevsky2012imagenet}. The representation problem gets more complicated
for vector data such as
point sets, polylines, polygons, and networks, which have more irregular spatial organization formats, because the concepts of location, distance, and direction among others do not have straightforward counterparts in existing 
NN
and it is not trivial to design NN operations (e.g., convolution) for irregularly structured
data \citep{valsesia2018learning}.

Early efforts perform data transformation operations to \textit{convert} the underlying spatial data into a format which can be handled by existing NN modules \citep{wang2019dynamic}. However, this conversion process often leads to information loss. For example, many early research about point cloud classification and segmentation first \textit{converted} 3D point clouds into volumetric representations (e.g., voxelized shapes) \citep{maturana2015voxnet,qi2016volumetric} or rendered them into 2D images \citep{su2015multi,qi2016volumetric}. Then they applied 3D or 2D CNN on these converted data representations for the classification or segmentation tasks. These practices have a major limitation -- choosing an appropriate spatial resolution for a volumetric representation is challenging \citep{qi2017pointnet}. A finer spatial resolution leads to data sparsity and higher computation cost while a coarser spatial resolution provides poor prediction results.

The reason for performing such data conversions 
is a lack of means to directly handle 
vector data in deep neural nets. An alternative approach is to encode these spatial data models directly. 
The first step towards such goal is to encode a point location into an embedding space such that these location embeddings can be easily used in the downstream NN modules. 
This is the idea of \textit{location encoding}.

Location encoding \citep{mac2019presence,mai2020multi,zhong2020reconstructing,mai2020se, gao2018learning,xu2018encoding,chu2019geo} refers to a 
NN-based encoding process which represents 
a point/location into a high dimensional vector/embedding such that this embedding can preserve different spatial information (e.g., distance, direction) and, at the same time, be \textit{learning-friendly} 

for downstream machine learning (ML) models such as neural nets 
and support vector machines (SVM).
 By \textit{learning friendly} we mean that the downstream model does not need to be very complex and does not require lots of training data to prevent model overfitting.
The encoding results are called location embeddings. And
the corresponding NN architecture is called
\textit{location encoder},  
which is a general-purpose model that
can be incorporated into different GeoAI models for different downstream tasks.

\begin{figure*}[t!]
	\centering 
	\includegraphics[width=\textwidth]{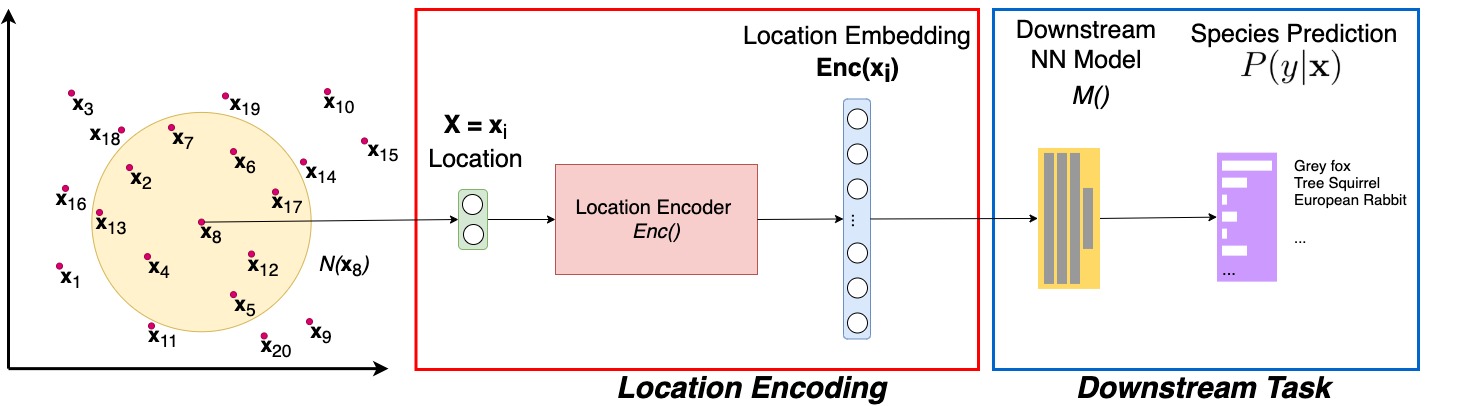}
	\caption[]{
	An illustration of location encoding. Here, we use \textit{location-based species classification} as an example of the downstream tasks. Those 20 points in 2D space represent species occurrence records. Each occurrence can be written as $\pt_i = (\bx_i, y_i)$ where $\bx_i$ indicates the 2D locations and $y_i$ indicates the corresponding species type, i.e., the ground truth label. $\neifunc(\bx_i)$ indicates the spatial neighborhood of $\bx_i$. A location encoder $\enc(\cdot)$ takes 2D location $\bx_i$ as its input and outputs a location embedding as a high dimensional vector. This embedding is further fed into a downstream NN model $M()$ for species prediction. The whole model architecture can be trained end-to-end in a supervised learning manner.
	}
	\label{fig:loc_enc}
	\vspace{-0.5cm}
\end{figure*}

Figure \ref{fig:loc_enc} is an illustration of location encoding. Here, we use location-based species classification as an example of the
downstream tasks which aims at predicting species $y$ based on a given location $\bx$. 
The training objective is to learn the conditional distribution $P(y|\bx)$, i.e., the probability of observing $y$ given 
$\bx$, which is highly non-linear. 
The idea of location encoding can be understood as a feature decomposition process which decomposes location $\bx$ (e.g.,  a two-dimensional vector of latitude and longitude) into a learning-friendly high dimensional vector (e.g., a vector with 100 dimensions), such that the highly non-linear distribution $P(y|\bx)$ can be learned with a relatively simple learner such as a linear SVM or a shallow 
NN model $M()$. The key benefits of such an architecture are to require less training data with simpler learners, and the possibility to leverage unsupervised training to better learn the location representations.

{Recently, the effectiveness of location encoding has been demonstrated in multiple GeoAI tasks including geo-aware image classification \citep{yin2019gps2vec,chu2019geo,mac2019presence,mai2020multi}, POI classification \citep{mai2020multi}, place annotation \citep{yin2019gps2vec}, trajectory prediction \citep{xu2018encoding,yin2019gps2vec}, location privacy protection \citep{rao2020lstm}, geographic question answering \citep{mai2020se}, 
3D protein distribution reconstruction \citep{zhong2020reconstructing},  point cloud classification and segmentation \citep{qi2017pointnet,qi2017pointnet++,li2018pointcnn}, and so on.
Despite these successful stories,
there is still a lack of a systematic review on such a topic. 
This paper fills this gap by providing a comparative survey on different location encoding models.
We give a general conceptual formulation framework which unifies almost all existing location encoding methods.}

It is worth mentioning that the location encoding discussed in this work is different from the traditional location encoding systems (i.e., geocoding systems)\footnote{\tiny{\url{https://gogeomatics.ca/location-encoding-systems-could-geographic-coordinates-be-replaced-and-at-what-cost/}}} which convert geographic coordinates into codes using an encoding scheme such as Geohash or codes for partition tiles such as Open Location Code and what3words. These traditional encoding systems are designed to support navigation and spatial indexing, while the neural location encoders we present here are used to support downstream ML models.

The contributions of our work are as follows:
	\begin{enumerate}
		\item Although there are multiple existing works on location encoding, 
		the necessity to design such a model is not clear. In this work, we formally define the location encoding problem and discuss the necessity from a machine learning perspective. 
		
		\item We conduct a systematic review on existing location encoding research. 
		A detailed classification system for location encoders is provided and all models are reformulated under a unified framework.
		This allows us to identify the commonalities and differences among different location encoding models. As far as we know, this is the first review on such a topic.

		\item We extend the idea of location encoding to the broader topic of 
		encoding different types of spatial data (e.g., polylines, polygons, graphs, and rasters). The possible solutions and challenges are discussed.
		
		\item To emphasize the general applicability of location encoding, we discuss its potential applications in different geoscience domains.
		 We hope these discussions can open up new areas of research.
	\end{enumerate}

The rest of this paper is structured as follows. We first introduce a formal definition of location encoding in Section \ref{sec:def}. Then, in Section \ref{sec:reason}, we discuss the necessity of 
location encoding. 
Next, we provide a general framework for understanding the current landscape of location encoding research and survey a collection of representative work in Section \ref{sec:review}. 
In Section \ref{sec:complex-geo}, we discuss how to apply location encoding for different types of spatial data. Finally, we conclude our work and discuss future research directions in Section \ref{sec:future}.

\section{Definitions}
\label{sec:def}

\begin{definition}[Location Encoding]
	\label{def:locenc}
	Given a set of points $\mP=\{\pt_i\}$, e.g., the locations of sensors, species occurrences, and so on, 
	where each point (e.g., an air quality sensor) $p_i=(\bx_i, \bv_i)$ is associated with a location $\bx_i \in \Real^{\spdim}$ (e.g., a sensor's location) in $\spdim$-D space ($\spdim = 2,3$) and attributes $\bv_i \in \Real^{\fedim}$ (e.g., air quality measurements). 
	We define the location encoder as a function $\enc^{(\mP, \theta)}(\bx): \Real^{\spdim} \to \Real^\embdim$ ($\spdim \ll \embdim$), 
	which is parameterized by $\theta$ and maps any coordinate $\bx$ in space to a vector representation of $\embdim$ dimension. 
	This process is called \textit{location encoding} and the results are called \textit{location embeddings}.

\end{definition}

Here, $\enc^{(\mP, \theta)}(\bx)$ indicates that the encoding result of $\bx$ may also depend on other locations in $\mP$. When $\enc^{(\mP, \theta)}(\bx)$ is independent of other points in $\mP$, we can simplify it to $\enc^{(\theta)}(\bx)$. 
Note that sometimes, the input of the location encoder can be both locations and attributes, i.e., $\enc^{(\mP,\theta)}(\bx, \bv): \Real^{\spdim+\fedim} \to \Real^\embdim$.

Figure \ref{fig:loc_enc} illustrates the idea of location encoding in Definition \ref{def:locenc}. The 20 2D points serve as an example of $\mP=\{\pt_i\}$ with $\spdim = 2$ and $n =  | \mP | = 20$. Note that $\enc^{(\mP, \theta)}(\bx)$ can not only be used to encode global location $\bx$, but also be utilized to encode the spatial relation between two locations, i.e., the spatial affinity vector $\Delta_{AB} = \bx_A - \bx_B$.

One question we may ask is \textit{whether a location encoder can preserve spatial information such as distance and direction information after the encoding process}.
From a spatial information preservation perspective, there are two properties we expect a location encoder to have: distance preservation and direction awareness.

\begin{property}[Distance Preservation]
	\label{prop:dist}
	The distance preservation property requires \textit{two nearby locations to have similar location embeddings. }
	More concretely, given any pair of location $(\bx_A, \bx_B)$, the inner product/similarity between their resulting location embeddings, i.e,  $\langle \enc^{(\mP, \theta)}(\bx_A), \enc^{(\mP, \theta)}(\bx_B) \rangle$ monotonically decreases when the distance\footnote{This can be Euclidean distance, manhattan distance, geodesic distance, great circle distance, and so on.}  between $\bx_A$ and $\bx_B$, i.e., $\parallel \bx_A - \bx_B \parallel$, increases. 
\end{property}

\begin{figure*}[t!]
	\centering 
	\includegraphics[width=0.35\textwidth]{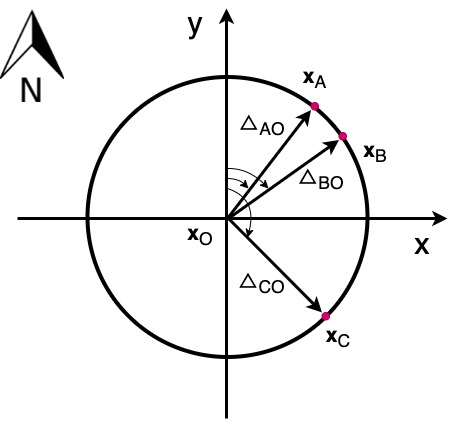}
	\caption[]%
	{An illustration of the direction preservation property of location encoding. 
	}
	\label{fig:direct_prop}
	\vspace{-0.5cm}
\end{figure*}

Property \ref{prop:dist} can be seen as a reflection of Tobler's First Law of Geography (TFL) \citep{tobler1970computer} in location encoding. 
The requirement of distance preservation has been adopted by multiple existing location encoding works. For example, \citet{gao2018learning} proposed a learnable location representation model $v(\bx)$ which consists of three sub-models: vector matrix multiplication, magnified local isometry, and global adjacency kernel. The global adjacency kernel sub-model assumes 
$\langle v(\bx_a),v(\bx_b) \rangle = (Kd)f(\parallel \bx_a - \bx_b \parallel)$, where $K$, $d$ are predefined constants and $f(r)$ is the adjacency kernel that decreases monotonically as $r$ increases. 
It can be seen that this sub-model directly satisfies Property \ref{prop:dist}. 
\citet{mai2020multi} also showed that their proposed multi-scale location encoder has a similar distance preservation property.

\begin{property}[Direction Awareness]
	\label{prop:direct}

		\textit{Locations that point into similar directions have more similar (relative) location embeddings than those who point into very different directions.} 
		More concretely, as shown in Figure \ref{fig:direct_prop}, given $\bx_O$ as the reference point and $y$ axis as the global north direction, $\bx_A$, $\bx_B$, and $\bx_C$ are on the same circle centered at $\bx_O$ and therefore share the same distance to $\bx_O$. 
		The relative spatial relation between $\bx_A$ and $\bx_O$ is defined as $\Delta_{AO} = \bx_A - \bx_O$. The direction of $\Delta_{AO}$, i.e., $\angle_{AO}$ is defined as the clockwise angle between $y$ axis and $\Delta_{AO}$. The same logic applies to $\bx_B$ and $\bx_C$.
		We say a location encoder is direction aware if it satisfies the following property: the inner product  $\langle \enc^{(\mP, \theta)}(\Delta_{AO}), \enc^{(\mP, \theta)}(\Delta_{BO}) \rangle > \langle \enc^{(\mP, \theta)}(\Delta_{AO}), \enc^{(\mP, \theta)}(\Delta_{CO}) \rangle$ if $|\angle_{AO} - \angle_{BO}|  < |\angle_{AO} - \angle_{CO}| $. 
\end{property}

Property \ref{prop:direct} is a reflection of the Generalized First Law of Geography \citep{zhu2019making} which includes direction into the consideration of similarities. 
In this paper, we call a location encoder an \textit{isotropic} location encoder if it only preserves the spatial proximity but ignores the variance of location embeddings when direction changes. We need to develop direction-aware, so-called anisotropic location encoders when the isotropicity assumption can not be held anymore.
In fact, in normal spatial analysis, isotropicity is the ``default'' assumption in most of the time. Although anisotropic versions of many geospatial analysis techniques have been developed such as directional kriging \citep{te2005mapping}, anisotropic clustering \citep{mai2018adcn}, direction remains on the level of an afterthought \citep{zhu2019making}. A similar situation can be seen in the current location encoding research, or GeoAI research in general. Compared with studies on distance preserved location encoders, there has been much less work on how to make a location encoder direction aware. \citet{mai2020multi} empirically showed that their multi-scale location encoder as well as many baseline models are direction-aware. However, this is just a by-product from their visualization analysis of the response maps of these location encoders. No theoretical proof is provided. As far as we know, there is no existing research that aims at developing a direction-aware location encoder deliberately.

Besides the above two spatial information preservation properties, Location Encoder $\enc^{(\mP, \theta)}()$ should also satisfy the following two properties to ensure its generalizability.

\begin{property}[Inductive Learning Method]
	\label{prop:inductive}
		Location Encoder $\enc^{(\mP, \theta)}()$ is an inductive learning method, i.e., the pretrained location encoder can be utilized to encode any location without retraining even if it does not appear in the training set. 
\end{property}

Property  \ref{prop:inductive} makes $\enc^{(\mP, \theta)}()$ differ from many existing transductive-learning-based location representation learning methods such as Location2Vec \citep{zhu2019location2vec}, POI2Vec \citep{feng2017poi2vec}, and \citet{kejriwal2017neural}. For example, \citet{kejriwal2017neural} converted a set of GeoNames locations into a k-th nearest neighbor graph in which locations (i.e., nodes) are linked to nearby locations by distance-weighted edges. A random-walk-based graph embedding method is used to learn an embedding for each location. This method is essentially a transductive learning model: when new locations are added to the training set, the graph is modified and the whole model has to be retrained to obtain the embeddings of new locations.

\begin{property}[Task Independence]
	\label{prop:task-indep}
	Location Encoder $\enc^{(\mP, \theta)}()$ should be task-independent or so called task-agnostic, i.e., the same model architecture can be used in different downstream tasks without any modification. 
\end{property}

Property  \ref{prop:task-indep} also differentiate $\enc^{(\mP, \theta)}()$ from some existing task-dependent location representation approaches. For instance, both Location2Vec \citep{zhu2019location2vec} and POI2Vec \citep{feng2017poi2vec} learned the embeddings of locations (e.g., cell stations, POIs) based on trajectories by adopting a Word2Vec-style training objective. This kind of location representation cannot be easily transferred to other tasks beyond human mobility. Similarly, \citet{gao2020representational} discretized the study area into $N \times N$ lattice and learned the embedding of each location by letting a Long Short Term Memory (LSTM) based artificial agent navigate through the study area. The learned location embeddings are used for simulation purpose but not for other geospatial tasks.

\begin{property}[Parametric Model]
	\label{prop:parametric}
    A \textit{parametric model} is a learning model with a finite set of parameters $\theta$. 
\end{property}

A parametric model is not very flexible, but the model complexity is bounded. In contrast, a \textit{non-parametric model} assumes that the data distribution cannot be defined with a finite set of parameters $\theta$ and the size of parameters $\theta$ can grow as the amount of data grows \mbox{\citep{russell2002artificial}}. So a non-parametric model is more flexible while the model complexity is unbounded.

Although Property \ref{prop:dist}, \ref{prop:direct}, and \ref{prop:parametric} are expected for location encoders, not all models we will discuss in Section \ref{sec:review} have these properties. See Table \ref{tab:review} for the detailed comparison. However, all location encoders discussed in Section \ref{sec:review} satisfy Property  \ref{prop:inductive} and \ref{prop:task-indep}. So we will not discuss these two properties separately for each model.

For simplicity, we will use $\enc()$ and $\enc^{(\mP)}()$ to indicate $\enc^{(\theta)}()$ and $\enc^{(\mP, \theta)}()$.

\section{The necessity of location encoding for GeoAI}
\label{sec:reason}
In this section we motivate the need to embed a location $\bx \in \Real^{\spdim}$ ($\spdim = 2,3$) into a high dimensional vector $\enc^{(\mP, \theta)}(\bx) \in \Real^\embdim $, which may seem counter-intuitive at first. 
{We mainly address this issue from a machine learning perspective.}

A key concept in statistical machine learning is bias-variance trade-off \citep{hastie2009}. On the one hand, when a learning system is required to pick one hypothesis out of a large hypothesis space  (e.g., deciding the parameters of a large 24 layer neural networks), \mai{it is flexible enough to approximate almost any non-linear distribution (low bias). However,} it needs a lot of training data to prevent overfitting.
This is called the low bias high variance situation.
On the other hand, when the hypothesis space is restricted (e.g., linear regression or single layer neural nets) the system has little chance to over fit, but might be ill-suited to model the underlying distribution 
and result in low performance in both training and testing sets \mai{(high bias)}.
This situation is called low variance high bias.
For many applications the data distribution is complex and \mai{highly non-linear.} We may not have enough domain knowledge to design good models with low variance (the effective model complexity) and low bias (the model data mismatch) at the same time. \mai{Moreover, we might want to avoid adopting too much domain knowledge into the model design which will make the resulting model task specific.} 
For example, the distribution of plant species (\mai{such as $P(y|\bx)$ in Figure \ref{fig:loc_enc}}) may be highly irregular influenced by several geospatial factors and interactions among species \citep{mac2019presence}.
Kernel (smoothing) methods (e.g., Radial Basis Function (RBF)) and neural networks (e.g., feed-forward nets) are  two types of most successful models which require very little domain knowledge for model design. 
They both have well established ways of controlling the effective model complexity. The kernel methods are more suited to low dimensional input data -- modeling highly non-linear distributions with little model complexity. \mai{However, they need to store the kernels during inference time which is not memory efficient.} Neural networks have more representation power \mai{which means a deep network can approximate very complex functions with no bias}, while requiring more domain knowledge for model design to achieve lower variance and bias.

\begin{figure*}[t!]
	\centering 
	\includegraphics[width=1.0\textwidth]{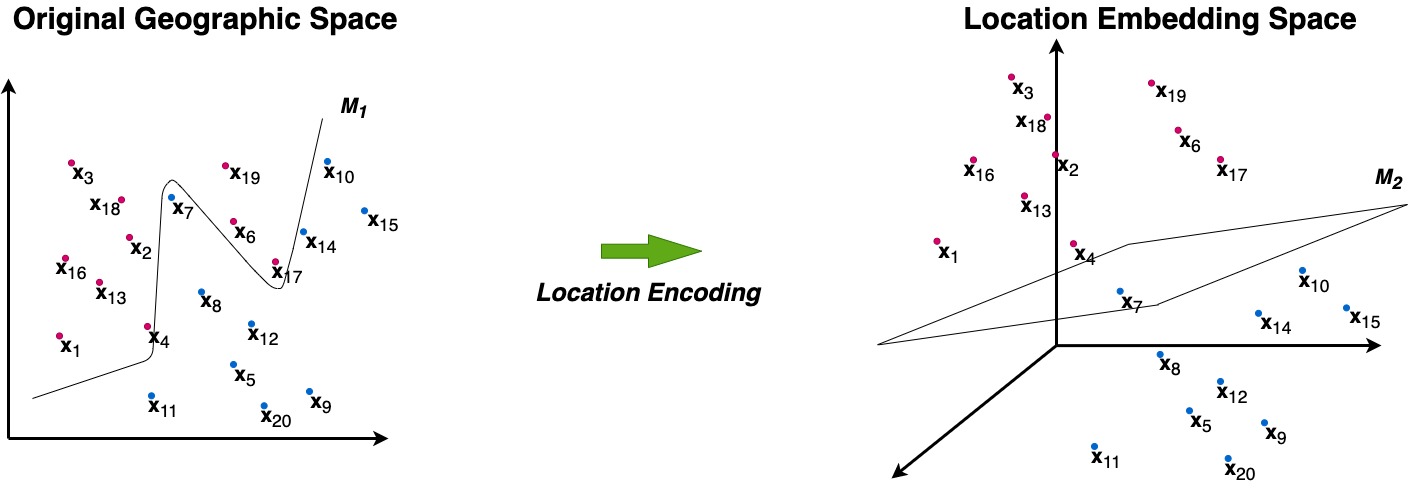}
	\caption[]%
	{
		An illustration of how location encoding can help to produce learning-friendly representations of geographic locations for downstream models. 
		We use the same 20 points in Figure \ref{fig:loc_enc} as an example of $\mP=\{\pt_i\}$. The red and blue points indicate they belong to two different classes.
		$M_1$ and $M_2$ are the illustrations of the trained binary classifiers in the original geographic space and the location embedding space. 
	}
	\label{fig:learn_friendly}
\end{figure*}

From a statistical machine learning perspective, the main purposes of location encoding is to produce 
{\it learning friendly representations of geographic locations} for downstream models such as SVM and neural networks.
By learning friendly  we mean that the downstream model does not need to be very complex and require large training samples.
For example,
 the location encoding process may 
perform a feature decomposition  ($\bx \in \Real^{\spdim} \to \enc^{(\mP, \theta)}(\bx) \in \Real^\embdim $, where $\spdim < \embdim$) 
so that the distribution we want to model such as $P(y|\bx)$ in Figure \ref{fig:loc_enc} becomes linear in the decomposed feature space, and a simple linear model can be applied. 
Figure \ref{fig:learn_friendly} illustrates this idea by using a simple binary classification task. If we use the original geographic coordinates $\bx$ as the input features to train the binary classifier, the resulting classifier $M_1$ will be a complex and nonlinear function which is prone to overfitting as shown in the left of Figure \ref{fig:learn_friendly}. After the location encoding process, the geographic coordinates feature is decomposed so that a simple linear model $M_2$ can be used as the binary classifier\footnote{The dimensionality of the location embedding space will be larger (e.g., 32 or 128); we use 3D here for ease of illustration.}.

\begin{table}[ht!]
	\caption{Overview of location encoding approaches. 
		Single point  location encoders $\enc(\bx)$ and {aggregation} location encoders $\enc^{(\mP)}(\bx)$
		 are further classified 
		based on either $\psenc(\bx)$ or $\neifunc(\bx)$ (see Figure \ref{fig:loc_enc}).
		(M) indicates multi-scale representation.
		$^*$ indicates a generalized version of the original model cited.
		We consider multiple criteria of location encoders:
		1) $\spdim$: The spatial dimension of $\mP$;
		2) Parametric: Is the location encoder a parametric model (Yes) or non-parametric model (No)?
		3) Mul.S.: Does the location encoding adopt a multi-scale approach?
		4) Dist.P.: Does this location encoder preserve distance (Property \ref{prop:dist})?
		5) Dir.A.: Is this location encoder aware of direction (Property \ref{prop:direct})?
		For Dist.P. and Dir.A.
		``Yes'' or ``No'' indicates whether the  property can be proved empirically (for example by using the response maps of trained location encoders \cite{mai2020multi}). 
		``-'' indicates that the  property is unknown. 
		``Yes+'' indicates that the property was shown both theoretically and empirically. 
	}
	\setlength{\tabcolsep}{2.5pt}
	\label{tab:review}
	\centering
	\scriptsize 
	\begin{tabular}{p{1.3cm}|p{1.9cm}|p{4.7cm}|c|c|c|c|c}
		\Xhline{4\arrayrulewidth}
		Encoder                                 & $\psenc(\bx)$                   & Model                                                                                 & $\spdim$ & Parametric & Mul.S. & Dist.P. & Dir.A. \\   \hline
		\multirow{7}{*}{$\enc(\bx)$}          & \multirow{2}{*}{Discretization} & \onehot \citep{tang2015improving}                     & 2,3      & Yes        & No          & No          & -              \\
		&                                 & \tile \citep{mai2020multi}                            & 2,3      & Yes        & No          & No          & -              \\  \cline{2-8}
		& Direct                          & \direct \citep{xu2018encoding,chu2019geo,rao2020lstm} & 2,3      & Yes        & No          & Yes         & Yes            \\   \cline{2-8}
		& Sinusoidal                      & \aodha \citep{mac2019presence}                        & 2        & Yes        & No          & Yes         & -              \\    \cline{2-8}
		& \multirow{3}{*}{Sinusoidal (M)} & \protein \citep{zhong2020reconstructing}              & 2,3      & Yes        & Yes         & -           & -              \\
		&                                 & \theory/Space2Vec \citep{mai2020multi}           & 2        & Yes        & Yes         & Yes+        & Yes            \\
		&                                 & \grid/Space2Vec \citep{mai2020multi}             & 2,3      & Yes        & Yes         & Yes         & -              \\   \Xhline{4\arrayrulewidth}
		Encoder                                 & $\neifunc(\bx)$                 & Model                                                                                 & $\spdim$ & Parametric & Mul.S. & Dist.Pr. & Dir.A. \\   \hline
		\multirow{10}{*}{$\enc^{(\mP)}(\bx)$} & \multirow{3}{*}{Kernel}         & \gpsvec \citep{yin2019gps2vec}                        & 2        & Yes        & No          &  -         & -              \\
		&                                 & \rbf \citep{mai2020multi}                             & 2,3      & Yes/No     & No          & Yes         & No             \\
		&                                 & Adapted kernel$^*$ \citep{berg2014birdsnap}                          & 2,3      & Yes/No     & No          & -           & -              \\   \cline{2-8}
		& Global                          & PointNet \citep{qi2017pointnet}                                      & 2,3      & Yes        & No          & -           & -              \\   \cline{2-8}
		& \multirow{3}{*}{Local}  & VoxelNet \citep{zhou2018voxelnet}                                    & 2,3      & Yes        & No          & -           & -              \\   
		&                          & \sagatmodel~ \citep{mai2020multi}                                    & 2,3      & Yes/No        & Yes/No          & Yes/No          & Yes/No              \\
		&                          & \dgcnnmodel~ \citep{wang2019dynamic}                                    & 2,3      & Yes        & No          & -          & -              \\ \cline{2-8}
		& \multirow{3}{*}{Hierarchical}    & PointNet++ \citep{qi2017pointnet++}                         & 2,3      & Yes        & Yes         & -           & -              \\
		&                                 & PointCNN \citep{li2018pointcnn}                                      & 2,3      & Yes        & Yes         & -           & -              \\
		&                                 & GraphCNN \citep{valsesia2018learning}                      & 2,3      & Yes        & Yes         & -           & -     \\        
		\Xhline{4\arrayrulewidth}     
	\end{tabular}
\end{table}

\section{A review of the current landscape of location encoding}
\label{sec:review}

In this section, we provide a comprehensive review of the existing location encoding techniques. 
Instead of enumerating every existing location encoding approach 
we organize our discussion in a top-down manner. 
We first classify location encoding models into different groups according to the input of location encoders and how they manipulate the spatial features. 
Firstly, according to the input, we can classify the existing location encoders into two categories: \textit{single point location encoder} $\enc(\bx)$ and \textit{aggregation location encoder} $\enc^{(\mP)}(\bx)$. 
$\enc(\bx)$ only considers the current point's location
while $\enc^{(\mP)}(\bx)$ additionally considers points in its neighborhood $\neifunc(\bx) \subseteq \mP$.
Then, in Section \ref{sec:single-pt}, $\enc(\bx)$ is further classified into sub-categories based on the type of positional encoder $\psenc(\bx)$.
Next, in Section \ref{sec:agg}, $\enc^{(\mP)}(\bx)$ is classified based on the used neighborhood $\neifunc(\bx)$.
The survey result is summarized in Table \ref{tab:review}. 
Finally, the comparison among different models will be discussed in Section \ref{sec:compare}.

\subsection{Single point location encoder  \texorpdfstring{$\enc(\bx)$}{}   }  
\label{sec:single-pt}

Interestingly, most existing single point location encoders, i.e., $\enc(\bx)$, 
\citep{tang2015improving,gao2018learning,xu2018encoding,chu2019geo,mac2019presence,mai2020multi,zhong2020reconstructing,rao2020lstm} share a similar structure:
\begin{align}
	\enc(\bx) = \pemlp(\psenc(\bx)),
	\label{equ:locenc}
\end{align}
Here, $\pemlp(\cdot): \Real^{\psdim} \to \Real^\embdim$ is a learnable neural network component which maps the input position embedding $\psenc(\bx) \in \Real^{\psdim}$ into the location embedding $\enc(\bx) \in \Real^\embdim$. A common practice is to define $\pemlp(\cdot)$ as a multi-layer perceptron, while \citet{mac2019presence} adopted a more complex $\pemlp(\cdot)$ which includes an initial fully connected layer, followed by a series of residual blocks. 
The purpose of $\pemlp(\cdot)$ is to provide a learnable component for the location encoder, which captures the complex interaction between input locations and target labels.

$\psenc(\cdot)$ is the most important component which distinguishes different $\enc(\bx)$. 
Usually, $\psenc(\cdot)$ is a \textit{deterministic} function which 
transforms location $\bx$ into a $\psdim$-dimension vector, so-called position embedding. 
The purpose of $\psenc(\cdot)$ is to do location feature normalization \citep{chu2019geo,mac2019presence,rao2020lstm} and/or feature decomposition \citep{mai2020multi,zhong2020reconstructing} so that the output $\psenc(\bx)$ is more learning-friendly for $\pemlp(\cdot)$. In Table \ref{tab:review} we further classify different $\enc(\bx)$ into four sub-categories based on their $\psenc(\cdot)$: discretization-based, direct, sinusoidal, and sinusoidal multi-scale location encoder. Each of them will be discussed in detail below.

\subsubsection{Discretization-based location encoder} \label{sec:discretization}
The early pioneers \citep{tang2015improving} argued that 
GPS coordinates are rather precise 
location indicators which are difficult to use by a classifier. So instead of using the 
coordinates, they discretized the whole study area into grid tiles and indicates each point by the corresponding grid that it falls into.

\begin{definition}[Discretization-based Location Encoder]
	\label{def:discretize}
	A discretization-based location encoder divides the study area (e.g., the earth surface) into regular area units such as grids, hexagons, or triangles - 
	$\enc_{\discretize}(\bx) = \pemlp(\psenc_{\discretize}(\bx))$ where $\psenc_{\discretize}(\cdot)$ is usually a tile lookup function which maps $\bx$ to a one hot vector that indicates the corresponding tile id it falls into. 
\end{definition}

\textbf{\onehot}. 
Early work in location encoding does not really have learnable component specific to the location encoder.
For example, \citet{tang2015improving} divided the study area (the contiguous United States) into $M$ rectangle grids. 
Given a location $\bx$ (the geotag of an image), 
$\psenc_{\onehot}(\bx) \in \Real^{M}$ is a one hot vector to indicate which grid $\bx$ falls
into and $\pemlp(\cdot)$ as an identity function, i.e., $\enc_{\onehot}(\bx) = \pemlp(\psenc_{\onehot}(\bx)) = \psenc_{\onehot}(\bx)$.

\textbf{\tile}. Later \citet{mai2020multi} introduced $\tile$ as one of $\enc_{\discretize}(\bx)$ which
uses a trainable location embedding matrix 
as $\pemlp(\cdot)$. This
makes it possible for the model to benefit from unsupervised training.

Although $\enc_{\discretize}(\bx)$ shows promising results on tasks such as geo-aware image classification, it has several inherent limitations: 
1) Each tile embedding are trained separately and spatial dependency is ignored, i.e., they do not have the distance preservation property;  
2) They have only one fixed spatial scale which can not effectively handle points with 
varied density;
3) Choosing the correct discretization is very challenging \citep{openshaw1981modifiable,fotheringham1991modifiable}, and incorrect choices will significantly affect the model's performance and efficiency \citep{lechner2012investigating}.

There are some possible solutions for these problems. For problem 1  we can add a regularization term in the 
loss function to make nearby tile embeddings have higher cosine similarity.  
For problem 2 and 3
one can adopt an adaptive discretization \citep{weyand2016planet} or multi-level discretization strategy as \citet{kulkarni2020spatial} did 
which uses deeper levels (smaller tiles) for higher point density areas and shallower levels (larger tiles) for sparse areas. However, finer spatial resolution or multi-level discretization means more tiles and more learnable parameters which can easily lead to overfitting.

\subsubsection{Direct location encoder} \label{sec:direct}

Recently, researchers adopted a rather simple approach {by directly applying neural networks to (normalized) coordinates} 
\citep{xu2018encoding,chu2019geo,rao2020lstm}.

\begin{definition}[Direct Location Encoder]
	\label{def:direct}
	A direct location encoder is defined as $\enc_{\direct}(\bx) = \pemlp(\psenc_{\direct}(\bx))$ where $\psenc_{\direct}(\bx)$ is usually a function to normalize or standardize the input location feature $\bx$ and $\pemlp(\cdot)$ is a multi-layer perceptron.
\end{definition}

\textbf{\direct}. 
There are many slight variations of $\enc_{\direct}(\bx)$ models. 
\citet{chu2019geo} took the input longitude and latitude  (i.e., $\bx = [\lon, \lat]^T$ of a image) and normalized them to range $[-1, 1)$ by dividing them with constant values. 
Similarly, in trajectory synthesis study, \citet{rao2020lstm} deployed $\psenc_{\direct}(\bx)$ which standardized each trajectory point $\bx = [\lon, \lat]^T$ by using the centroid of all trajectory points. 
In order to perform pedestrian 
trajectory prediction, \citet{xu2018encoding} also designed a simple location encoder whose $\psenc_{\direct}(\bx)$ normalizes $\bx$ to $[0,1]$. 
Without a feature decomposition step, these models often fail to capture the fine details of data distributions, and have worse prediction accuracy than $\tile$ on specific tasks.

\subsubsection{Sinusoidal location encoder} \label{sec:sinusoidal}

\begin{definition}[Sinusoidal Location Encoder]
	\label{def:sinusoidal}
	A sinusoidal location encoder is defined as $\enc_{\sinusoidal}(\bx) = \pemlp(\psenc_{\sinusoidal}(\bx))$ where $\psenc_{\sinusoidal}(\bx)$ is a deterministic function which processes $\bx$ with sinusoidal functions, e.g., $\sin()$, 
	after location feature normalization.
\end{definition}

\textbf{\aodha}. 
\citet{mac2019presence} developed $\enc_{\aodha}(\bx) =  \pemlp(\psenc_{\sinusoidal}(\bx))$ which uses sinusoidal functions to wrap the geographic coordinates. 
In Equation \ref{equ:wrap}, longitude $\lon$ and latitude $\lat$ are first normalized into range $[-1, 1]$ by dividing by $180^{\circ}$ and $90^{\circ}$ accordingly and then are fed into $\sin(\pi x)$ and $\cos(\pi x)$ functions.
\begin{align}
	\psenc_{\aodha}(\bx) & = [\sin(\pi \dfrac{\lon}{180^{\circ}}), \cos(\pi \dfrac{\lon}{180^{\circ}}), \sin(\pi \dfrac{\lat}{90^{\circ}}), \cos(\pi \dfrac{\lat}{90^{\circ}})], \;
	where \; \bx = (\lon, \lat)
	\label{equ:wrap}
\end{align}

The purpose to use sinusoidal functions is
to wrap geographic coordinates around the world \cite{mac2019presence}. 
This ensures that longitude $\lon_1 = -180^{\circ}$ and $\lon_2 = 180^{\circ}$ have the same encoding results.
However, applying this encoding strategy to latitudes is problematic. 
As for $\lat_1 = -90^{\circ}$ and $\lat_2 = 90^{\circ}$, 
i.e., the South pole and North pole, 
they would
have identical encoding results which is problematic. 
Moreover, even if we fix this problem, $\aodha$
 is still not a spherical distance-preserved location encoder.

\subsubsection{Sinusoidal multi-scale location encoder} \label{sec:multiscale}

One limitation of all location encoders we discussed so far is that they can not handle non-uniform point density \citep{qi2017pointnet} or mixtures of distributions with very different characteristics \citep{mai2020multi}. 
For example, given a set of POIs, some types tend to cluster together such as night clubs, women's clothing, restaurants while other POI types are rather evenly distributed such as post offices, schools, and fire stations. 
Similarly, as for species occurrences, some species herd together such as wildebeests and zebras while the individuals of other species tend to walk alone and protect their own territory such as tigers and bears. This will result in different spatial distribution patterns of these species occurrences.
In order to jointly model these spatial distributions, 
we need an encoding method which supports multi-scale representations.

Inspired by the position encoding architecture in Transformer \citep{vaswani2017attention}, researchers developed multi-scale location encoders by using sinusoidal functions with different frequencies \citep{mai2020multi,zhong2020reconstructing}.

\begin{definition}[Sinusoidal Multi-Scale Location Encoder]
	\label{def:sin-multi}
	The sinusoidal multi-scale location encoder is defined as $\enc_{\sinmulti}(\bx) = \pemlp(\psenc_{\sinmulti}(\bx))$ where $\psenc_{\sinmulti}(\bx)$ decomposes $\bx$ into a multi-scale representation based on different sinusoidal functions with different frequencies:
	\begin{align}
		\psenc_{\sinmulti}(\bx) & =  [\psenc^{(\nscale)}_{0}(\bx);...;\psenc^{(\nscale)}_{\certainscale}(\bx);...;\psenc^{(\nscale)}_{\nscale-1}(\bx)]. 
		\label{equ:sin-multi}
	\end{align}
	Here $\nscale$ is the total number of scales. 
	$\psenc^{(\nscale)}_{\certainscale}(\bx)$ processes the location features with different sinusoidal functions whose frequency is determined by the scale $\certainscale$.
\end{definition}

\textbf{\protein}. \citet{zhong2020reconstructing} slightly changed the position encoding architecture in Transformer \citep{vaswani2017attention} and applied them in high dimension data points such as 3D Cartesian coordinates.

\begin{definition}[Transformer-based Location Encoder]
	\label{def:protein}
	The transformer-based location encoder $\enc_{\protein}(\bx) = \pemlp(\psenc_{\protein}(\bx))$ is following Definition \ref{def:sin-multi}.
	For each scale $\certainscale \in \{0,1,...,\nscale-1\}$, 
	$\psenc^{(\nscale)}_{\certainscale}(\bx) = \psenc^{\protein}_{\certainscale}(\bx)$ 
	is defined by Equation \ref{equ:protein}. Here, $\bx^{[l]}$ indicates the $l$th dimension of $\bx$. 
	\begin{subequations}
		\label{equ:protein}
		\begin{align}	
		\psenc^{\protein}_{\certainscale}(\bx) = & [\psenc^{\protein}_{\certainscale,1}(\bx);...;\psenc^{\protein}_{\certainscale,l}(\bx);...;\psenc^{\protein}_{\certainscale,\spdim}(\bx)], \label{equ:protein-1} \\
		where \; \psenc^{\protein}_{\certainscale,l}(\bx) & = [\cos(   \dfrac{2\pi\nscale \bx^{[l]}}{\nscale^{(\certainscale+1)/\nscale}});  \sin( \dfrac{2\pi\nscale \bx^{[l]}}{\nscale^{(\certainscale+1)/\nscale}})], \; \forall l = 1,2,...,\spdim.
		\label{equ:protein-2}
		\end{align}
	\end{subequations}
\end{definition}

\citet{zhong2020reconstructing} showed that $\enc_{\protein}(\bx)$ works well for noiseless data, but for noisy data they need to exclude the top 10\% highest frequency components (the smallest several $\certainscale$) in Equation \ref{equ:sin-multi}. 
This indicates the necessity of another parameter to control the smallest scale we consider in sinusoidal functions. That is the usage of $\minscale$ in $\theory$ and $\grid$ which we will discussed below.

\textbf{\theory}. Space2Vec \citep{mai2020multi} introduced $\theory$ as a 2D multi-scale location encoder by using sinusoidal functions with different frequencies. 

\begin{definition}[Theory Location Encoder]
	\label{def:theory}
	Let $\unit_1 = [1,0]^T, \unit_2 = [-1/2,\sqrt{3}/2]^T, \unit_3 = [-1/2,-\sqrt{3}/2]^T \in \Real^2$ be three unit vectors which are oriented $120^{\circ}$ apart from each other. $\minscale, \maxscale$ are the minimum and maximum grid scale, and $\scaleratio = \frac{\maxscale}{\minscale}$. 
	$\enc_{\theory}(\bx)$ is following Definition \ref{def:sin-multi} where in each scale $\certainscale \in \{0,1,...,\nscale-1\}$, $\psenc^{(\nscale)}_{\certainscale}(\bx) = \psenc^{\theory}_{\certainscale}(\bx)$ is defined in Equation  \ref{equ:theory}. Here, $\langle \cdot, \cdot \rangle$ indicates vector inner product.
	\begin{subequations}
		\label{equ:theory}
	\begin{align}
		\psenc^{\theory}_{\certainscale}(\bx) & = [\psenc^{\theory}_{\certainscale,1}(\bx);\psenc^{\theory}_{\certainscale,2}(\bx);\psenc^{\theory}_{\certainscale,3}(\bx)], \label{equ:theory-1} \\
		where \; \psenc^{\theory}_{\certainscale,j}(\bx) & = [
		\cos(\frac{\langle \bx, \unit_{j} \rangle}{ \minscale \cdot \scaleratio^{\certainscale/(\nscale-1)}}); 
		\sin(\frac{\langle \bx, \unit_{j} \rangle}{ \minscale \cdot \scaleratio^{\certainscale/(\nscale-1)}})], \; \forall j = 1,2,3.
		\label{equ:theory-2}
	\end{align}
	\end{subequations}
	
\end{definition}

Both $\enc_{\theory}(\cdot)$ and \citet{gao2018learning} are inspired by the grid cell research from neuroscience field \citep{hafting2005microstructure,blair2007scale,killian2012map,agarwal2015boon}. 
In fact, \citet{gao2018learning} inspired and laid the theoretical foundation of $\enc_{\theory}(\bx)$. As we discussed in Section \ref{sec:def}, \citet{gao2018learning} proposed a location representation model $v(\bx)$ which consists of three sub-models. They also proposed a complex-value-based location encoder $\theofuc(\bx)$ as an analytical solution for $v(\bx)$ which inspired $\enc_{\theory}(\cdot)$. More specifically,
given two location $\bx_a$, $\bx_b$, \citet{gao2018learning} proved that 
$\langle \theofuc(\bx_a), \theofuc(\bx_b) \rangle = 3(1 - \dfrac{\beta}{4}\parallel \bx_b - \bx_a \parallel^2)$ where $\beta = \parallel \unit_{j} \parallel^2_2 = 1$.
That means the inner products between their location embeddings 
increase when $\parallel \bx_b - \bx_a \parallel^2$ decrease, which satisfies Property \ref{prop:dist}. 
\citet{mai2020multi} showed that $\enc_{\theory}(\cdot)$ also satisfies Property \ref{prop:dist} both theoretically and empirically.

\textbf{\grid}. \textit{\grid} is another type of $\enc_{\sinmulti}(\bx)$
proposed by Space2Vec \citep{mai2020multi}. 

\begin{definition}[Grid Location Encoder]
	\label{def:grid}
	$\enc_{\grid}(\bx)$ follows Definition \ref{def:sin-multi} where at each scale $\certainscale \in \{0,1,...,\nscale-1\}$, $\psenc^{(\nscale)}_{\certainscale}(\bx) = \psenc^{\grid}_{\certainscale}(\bx)$ is defined by Equation \ref{equ:grid}. Here, $\minscale, \maxscale$ and $\scaleratio$ follow the same definition as Definition \ref{def:theory}. 
	\begin{subequations}
		\label{equ:grid}
	\begin{align}
		\psenc^{\grid}_{\certainscale}(\bx) & = [\psenc^{\grid}_{\certainscale,1}(\bx);...;\psenc^{\grid}_{\certainscale,l}(\bx);...;\psenc^{\grid}_{\certainscale,\spdim}(\bx)], 	\label{equ:grid-1} \\
		where \; \psenc^{\grid}_{s,l}(\bx) & =  [
		\cos(\frac{\bx^{[l]}}{ \minscale \cdot \scaleratio^{\certainscale/(\nscale-1)} }); 
		\sin(\frac{\bx^{[l]}}{ \minscale \cdot \scaleratio^{\certainscale/(\nscale-1)} })], \; \forall l = 1,2,..,\spdim.
	\label{equ:grid-2}
	\end{align}
	\end{subequations}
	
\end{definition}

\citet{mai2020multi} shows that for both $\theory$ and $\grid$, $\maxscale$ can be directly determined based on the size of the study area while $\minscale$ is the critical parameter which decides the highest spatial resolution $\enc(\bx)$ can handle. 

\subsubsection{Comparison among different $\enc(\bx)$}

Compared with $\discretize$ 
which yields identical embeddings for $\bx$ that fall into the same tile,  $\direct$ can distinguish nearby locations, 
i.e., $\enc_{\direct}(\bx_a) \neq \enc_{\direct}(\bx_b)$, if $ \bx_a \neq \bx_b $. 
\citet{mai2020multi} compared them in different tasks and found out that without an appropriate location feature normalization 
$\psenc_{\direct}(\bx)$, $\direct$ will show lower performance than $\tile$.
This indicates the importance of $\psenc_{\direct}(\bx)$.

One advantage of $\direct$ is its simple architecture with fewer hyperparameters. However, compared with $\psenc_{\sinusoidal}(\bx)$ and $\psenc_{\sinmulti}(\bx)$, $\psenc_{\direct}(\bx)$ is rather hard for $\pemlp(\cdot)$ to learn from and may produce over-generalized distributions.

Compared with $\protein$ and $\grid$, 
$\theory$ has a theoretical foundation to ensure Property \ref{prop:dist}. 
However, $\theory$ can only be applied to point sets in 2D space. 
In contrast, $\protein$  and $\grid$ lack a theoretical guarantee for Property \ref{prop:dist}
while they can be utilized for points in any $\spdim$-D space. $\protein$ and $\grid$ follow similar idea while $\grid$ has an additional parameter $\minscale$, which is more flexible for data sets with different characteristics.

\subsection{Aggregation location encoder $\enc^{(\mP)}(\bx)$}   \label{sec:agg}

\begin{definition}[Aggregation Location Encoder]
	\label{def:agg-locenc}
	The aggregation location encoder $\enc^{(\mP)}(\bx)$ jointly considers the location feature 
	$\bx$ and the aggregated features from the neighborhood of $\bx$, denoted as $\neifunc(\bx)$.  
	Inspired by the Graph Neural Network (GNN) framework \citep{xu2018powerful}, a generic model setup of $\enc^{(\mP)}(\bx)$ can be defined as 
\begin{subequations}
	\label{equ:aggenc}
	\begin{align}
	\locemb^{(0)}_{\bx} & = \enc(\bx), \label{equ:aggenc_1}\\
	\aggemb^{(\kthagglayer)}_{\bx} & = \aggfunc_{\bx_i \in \neifunc(\bx)}\{\locemb^{(\kthagglayer-1)}_{\bx_i}\} , \label{equ:aggenc_2}\\
	\locemb^{(\kthagglayer)}_{\bx} & = \comfunc(\locemb^{(\kthagglayer-1)}_{\bx}, \aggemb^{(\kthagglayer)}_{\bx}), \label{equ:aggenc_3}\\
	\enc^{(\mP)}(\bx)   & = \readoutfunc(\locemb^{(\numagglayer)}_{\bx}). \label{equ:aggenc_4}
	\end{align}
\end{subequations}
\end{definition}

It consists of $\numagglayer$ Location Encoding Aggregation (LEA) layers, which iteratively update the location representations. 
In Equation \ref{equ:aggenc_1}, the initial location embedding $\locemb^{(0)}_{\bx}$ can be computed based on any single point location encoder discussed in Section \ref{sec:single-pt}.
Each LEA layer constitutes one neighborhood aggregation operation 
$ \aggfunc_{\bx_i \in \neifunc(\bx)}\{\cdot\}$ 
(Equation \ref{equ:aggenc_2}) and a feature combination operation 
$\comfunc(\cdot, \cdot)$ (Equation \ref{equ:aggenc_3}).
$ \aggfunc_{\bx_i \in \neifunc(\bx)}\{\cdot\}$ 
aggregates the feature $\locemb^{(\kthagglayer-1)}_{\bx_i}$ of $\bx_i$ in 
$\neifunc(\bx)$ 
from the previous LEA layer
which can be seen as an analogy of the convolution operation of CNN on point sets. $ \aggfunc_{\bx_i \in \neifunc(\bx)}\{ \cdot \}$ can be element-wise max/min/mean pooling, 
sum, or any permutation invariant architectures \citep{zaheer2017deep,qi2017pointnet,velivckovic2018graph,mai2019contextual}. $\comfunc(\cdot, \cdot)$ 
combines the point-wise feature $\locemb^{(\kthagglayer-1)}_{\bx}$ with the aggregated features $\aggemb^{(\kthagglayer)}_{\bx}$ which can be vector concatenation, element-wise max, min or mean. 
In the last layer, a readout function $\readoutfunc(\cdot)$, which can be an identity function or a multi-layer perceptron, produces the final aggregated location embedding for $\bx$ (See Equation \ref{equ:aggenc_4}).

Equation \ref{equ:aggenc} can be treated as an analogy of the 
GNN framework \citep{battaglia2018relational,xu2018powerful,wu2020comprehensive} which is a general framework for different neural network architectures applied on graphs such as GCN 
\citep{kipf2016semi}, GraphSAGE \citep{hamilton2017inductive}, GG-NN 
\citep{li2015gated}, GAN 
\citep{velivckovic2018graph}, MPNN 
\citep{gilmer2017neural}, R-GCN \citep{schlichtkrull2018modeling}, CGA 
\citep{mai2019contextual}, TransGCN \citep{cai2019transgcn}, and so on.

$\neifunc(\bx)$ can be defined in different ways such as the top kth nearest locations \citep{apple2020kriging,mai2020multi}, locations within a buffer radius \citep{qi2017pointnet++}, or locations within the same voxel as $\bx$ \citep{zhou2018voxelnet}. 
According to the definition of $\neifunc(\bx)$, we classify $\enc^{(\mP)}(\bx)$ into different categories: kernel, global, local neighborhood, and hierarchical 
neighborhood aggregation location encoder 
as summarized in Table \ref{tab:review}. We  will  discuss each in detail in the following section. 

\subsubsection{Kernel-based location encoder} \label{sec:kernel}

A kernel-based location encoder needs two components: a predefined kernel function $\kernelfunc(\cdot, \cdot)$ and a set of kernel center points $\mQ=\{\pt_j\}$.
The selection of $\kernelfunc(\cdot, \cdot)$ depends on the nature of the dataset. The popular options are RBF kernels and Mercer kernels \citep{vapnik2013nature}. 
$\mQ$ can be equal to or a subset of the training point set, i.e., $\mQ \subseteq \mP$ 
or can be a predefined point set such as the centers of regular grids 
\citep{yin2019gps2vec}.

\begin{definition}[Kernel-Based Location Encoder]
	\label{def:kernel}
	Given a kernel function $\kernelfunc(\cdot, \cdot)$ and the kernel center point set $\mQ=\{\pt_j\}$, 
	we define the kernel-based location encoder as Equation \ref{equ:kernel} by following Definition \ref{def:agg-locenc} where $\numagglayer = 1$. 
	\begin{subequations}
		\label{equ:kernel}
		\begin{align}
			\locemb^{(0)}_{\bx} & = \bx, \label{equ:kernel_1} \\
			\aggemb^{(1)}_{\bx} & = \aggfunc^{(\kernel)}_{\pt_i \in \neifunc(\bx)}\{\locemb^{(0)}_{\bx_i}\} = \kerneldecomp(\bx) = [\kernelfunc(\bx, \bx_1);...;\kernelfunc(\bx, \bx_{|\mQ|})], \label{equ:kernel_2} \\
			\locemb^{(1)}_{\bx} & = \comfunc(\locemb^{(0)}_{\bx}, \aggemb^{(1)}_{\bx}) = 	\aggemb^{(1)}_{\bx}, \label{equ:kernel_3} \\
			\enc^{(\mP)}_{\kernel}(\bx)   & = \readoutfunc(\locemb^{(1)}_{\bx}) = \pemlp(\locemb^{(1)}_{\bx}).
			\label{equ:kernel_4}
		\end{align}
	\end{subequations}
	Here, each $\bx$ has a fixed ``neighborhood'' - $\mQ$, i.e., $\neifunc(\bx) = \mQ = \{\pt_j = (\bx_i)\}$. $[\cdot;\cdot]$ indicates vector concatenation. $| \mQ |$ indicates the total number of kernels and $\pemlp(\cdot)$ is a multi-layer perceptron.
\end{definition}

From a location feature decomposition perspective, $\enc^{(\mP)}_{\kernel}(\bx)$ can be seen as using kernel trick to decompose $\bx$ into multiple kernel features (see Section \ref{sec:reason}).
From a location encoding aggregation perspective, it can be seen as aggregating the kernel features derived from the interaction between $\bx$ and each kernel.
Since each location $\bx \in \mP$ shares the same $\neifunc(\bx) = \mQ$, $\aggfunc^{(\kernel)}_{\pt_i \in \neifunc(\bx)}\{\cdot\}$ does not need to be a permutation invariant function. Here we use a concatenation of the $| \mQ |$ kernel features of $\bx$.
Examples of $\enc^{(\mP)}_{\kernel}(\bx)$ include GPS2Vec \citep{yin2019gps2vec} and the $\rbf$ baseline used by Space2Vec \citep{mai2020multi}. The main difference among them are the definitions of $\mQ$ and $\kernelfunc(\cdot, \cdot)$.

\textbf{GPS2Vec}. GPS2Vec \citep{yin2019gps2vec} divided the Earth into multiple UTM zones and trained different $\enc^{(\mP)}_{\kernel}(\bx)$ separately. Each zone is further divided into $\numkernel$ grids whose centers are used as $\mQ$ and $\kernelfunc(\bx, \bx_j) = \exp(-\dfrac{\parallel \bx - \bx_j \parallel_2}{\regularkernel})$. Here $\parallel \cdot \parallel_2$ indicates L2 norm and $\regularkernel$ is a constant attenuation coefficient. We denote this model as $\enc^{(\mP)}_{\gpsvec}(\cdot)$.

\textbf{\rbf}. \citet{mai2020multi} proposed a RBF kernel based location encoder as one of their baselines, called $\rbf$. $\mQ$ are randomly sampled $\numkernel$ points from $\mP$, i.e., $\mQ \subseteq \mP$. 
The kernel function $\kernelfunc(\bx, \bx_j) = \exp{\big(-\dfrac{\parallel \bx  - \bx_j \parallel^2_{2}}{2\sigma^2}\big)}$ is a RBF kernel. 
We denote it as $\enc^{(\mP)}_{\rbf}(\cdot)$.

\textbf{Adapted kernel*} 
Instead of using constant kernel bandwidth, \citet{berg2014birdsnap} also proposed an idea of adaptive kernels to model the spatio-temporal distribution prior of bird species. They precomputed this prior as a fixed scale kernel density estimation (KDE) map. Here, instead of making a precomputed KDE map, we adopt this idea to design an adaptive kernel based location encoder $\enc_{\adpativekernel}(\bx)$ in which $\kernelfunc(\bx, \bx_j) = \exp(-\dfrac{\parallel \bx - \bx_j \parallel^2_2}{2\adpativekernelwidth(\bx)^2})$. $\adpativekernelwidth(\bx)$ is a kernel bandwidth function of location $\bx$, e.g., defining $\adpativekernelwidth(\bx)$
as the half of the distance from $\bx$ to its $K$th nearest neighbor. 
We use * to differentiate $\enc_{\adpativekernel}(\bx)$ from the original model proposed by \citet{berg2014birdsnap}.

Despite its simple design and the ability to handle non-linear distributions, $\enc_{\kernel}(\bx)$ has several shortcomings.
Firstly, $\enc_{\kernel}(\bx)$ has to memorize $\mQ$ at testing time which will affect memory efficiency. Secondly, since $\locemb^{(1)}_{\bx} \in \Real^{| \mQ |}$, the size of $\mQ$ directly decides the number of learnable parameters in $\pemlp(\cdot)$. This creates a performance-efficiency trade-off problem. 
When $| \mQ |$ is small, $\enc_{\kernel}(\bx)$ has less memory burden. $\pemlp(\cdot)$ also has fewer learnable parameters which need less training data and is less likely to over fit. However, $\mQ$ is rather distributed sparsely over the study area which affects the quality of the encoding results.
When $| \mQ |$ is rather large, the encoding results are more accurate, but we need a huge amount of memory to store $\mQ$ and $\pemlp(\cdot)$ needs more learning parameters. 
Moreover, the prediction of $\enc_{\kernel}(\bx)$ also depends on the distribution of $\mQ$ since it performs poorly on data sparse regions.

\subsubsection{Global aggregation location encoder}   \label{sec:global-agg}

The global aggregation location encoder  $\enc^{(\mP)}_{\globalagg}(\bx)$ defines $\neifunc(\bx)$ as all locations in $\mP$, i.e., $\neifunc_{\globalagg}(\bx) = \mP$.

\textbf{PointNet}. PointNet \citep{qi2017pointnet}
was originally proposed for 3D point cloud classification and segmentation. 
Its point cloud segmentation architecture 
can be formulated as a global aggregation location encoder 
as shown in Equation \ref{equ:pointnet-agg}. 
Equation \ref{equ:pointnet_1} first embeds each point into an initial embedding with a $\direct$-like single point location encoder $\enc_{\poinet}(\bx)$.
It normalizes the input 3D point feature $\bx = [x, y, z]^T \in \Real^3$ on to a unit sphere, denoted as $\psenc_{\poinet}(\cdot)$, before feeding them into $\pemlp(\cdot)$. 
$\pemlp(\cdot)$ consists of two affine transformations (parameterized as t-nets) $\transnet_{1}(\cdot)$ and $\transnet_{2}(\cdot)$, which are separated by a multi-layer perceptron $\mlp_{1}(\cdot)$. 
These affine transformations help to make the semantic labeling of a point cloud invariant to geometric transformations \citep{qi2017pointnet}.
Note that PointNet only has one 
LEA layer, i.e., $\numagglayer = 1$. 
$\aggfunc^{(\poinet)}_{\bx_i \in \neifunc(\bx)}\{\cdot\}$ 
is a multi-layer perceptron $\mlp_{2}(\cdot)$ for each $\bx_i \in \neifunc_{\poinet}(\bx)= \mP$ followed by an element-wise max pooling 
$\maxpool_{\bx_i \in \neifunc_{\poinet}(\bx)}\{\cdot\}$ (See Equation \ref{equ:pointnet_2}). 
$\comfunc(\cdot, \cdot)$ 
is a vector concatenation operation followed by a multi-layer perceptron $\mlp_{3}(\cdot)$ and the readout function 
$\readoutfunc(\cdot)$ is an identity function as shown in Equation \ref{equ:pointnet_3}, \ref{equ:pointnet_4}. 

\begin{subequations}
    \vspace{-0.5cm}
	\label{equ:pointnet-agg}
	\begin{align}
	\locemb^{(0)}_{\bx} & = \enc_{\poinet}(\bx) = \pemlp(\psenc_{\poinet}(\bx)) = \transnet_{2}(\mlp_{1}(\transnet_{1}(\psenc_{\poinet}(\bx))), \label{equ:pointnet_1}\\
	\aggemb^{(1)}_{\bx} & = \aggfunc^{(\poinet)}_{\bx_i \in \neifunc_{\poinet}(\bx)}\{\locemb^{(0)}_{\bx_i}\} = \maxpool_{\bx_i \in \neifunc_{\poinet}(\bx)}\{\mlp_{2}(\locemb^{(0)}_{\bx_i})\}, \label{equ:pointnet_2} \\
	\locemb^{(1)}_{\bx} & = \comfunc^{(\poinet)}(\locemb^{(0)}_{\bx}, \aggemb^{(1)}_{\bx}) = \mlp_{3}([\locemb^{(0)}_{\bx}; \aggemb^{(1)}_{\bx}]), \label{equ:pointnet_3}\\
	\enc^{(\mP)}_{\poinet}(\bx)&  = \readoutfunc^{(\poinet)}(\locemb^{(1)}_{\bx}) = \locemb^{(1)}_{\bx} . 
	\label{equ:pointnet_4} 	
	\end{align}
\end{subequations}

\subsubsection{Local aggregation location encoder}   \label{sec:single-scale-agg} 
The local aggregation location encoder $\enc^{(\mP)}_{\singleagg}(\bx)$ considers a 
local neighborhood $\neifunc(\bx)$ such as 
all locations within a buffer of $\bx$ with radius $\buffradius$, i.e.,
$\neifunc_{\buffer}(\bx) = \{\bx_i | \parallel \bx - \bx_i \parallel_2 \leq \buffradius, \; \forall \bx_i \in \mP \wedge \bx_i \neq \bx \}$.

\textbf{VoxelNet}.  \citet{zhou2018voxelnet} 
discretizes the 3D space into unit voxels. 
For any location $\bx$, its neighborhood is defined as 
$\neifunc_{\voxnet}(\bx) = \{\bx_i | \voxlocate(\bx) = \voxlocate(\bx_i), \; \forall \bx_i \in \mP\}$,
where $\voxlocate(\bx)$ is a voxel lookup function which returns the ID of the voxel in which $\bx$ falls into. 
\begin{subequations}
	\label{equ:voxelnet-agg}
	\begin{align}
	\locemb^{(0)}_{\bx} & = \enc_{\voxnet}(\bx) = \pemlp(\psenc_{\voxnet}(\bx)) = \psenc_{\voxnet}(\bx), \label{equ:voxelnet_1}\\
	\aggemb^{(\kthagglayer)}_{\bx} & = \aggfunc^{(\voxnet)}_{\bx_i \in \neifunc_{\voxnet}(\bx)}\{\locemb^{(\kthagglayer-1)}_{\bx_i}\} = \maxpool_{\bx_i \in \neifunc_{\voxnet}(\bx)}\{\fcn^{(\kthagglayer)}(\locemb^{(\kthagglayer-1)}_{\bx_i})\}, \label{equ:voxelnet_2} \\
	\locemb^{(\kthagglayer)}_{\bx} & = \comfunc^{(\voxnet)}(\locemb^{(\kthagglayer-1)}_{\bx}, \aggemb^{(\kthagglayer)}_{\bx}) = [\locemb^{(\kthagglayer-1)}_{\bx}; \aggemb^{(\kthagglayer)}_{\bx}], \label{equ:voxelnet_3}\\
	\enc^{(\mP)}_{\voxnet}(\bx)&  = \readoutfunc^{(\voxnet)}(\locemb^{(\numagglayer)}_{\bx}) = \locemb^{(\numagglayer)}_{\bx}.
	\label{equ:voxelnet_4} 	
	\end{align}
\end{subequations}
First, a $\direct$-like location encoder $\enc_{\voxnet}(\cdot)$ (Equation \ref{equ:voxelnet_1}) encodes $\bx = [x, y, z]^T$ into 
$\psenc_{\voxnet}(\bx) = [x, y, z, x - \bar{x}, y - \bar{y}, z - \bar{z}]^T$. Here $\pemlp(\cdot)$ is an identity function\footnote{In \citet{zhou2018voxelnet}, they encode each 3D point as  $[x, y, z, v, x - \bar{x}, y - \bar{y}, z - \bar{z}]^T$ where $v$ is an attribute of each point, 
e.g., received reflectance. 
Here we skip $v$ to focus on the location information. However, as we said in Definition \ref{def:locenc}, it is very easy to extend  $\enc^{(\mP, \theta)}(\bx)$ to $\enc^{(\mP,\theta)}(\bx, \bv)$.}, and
$(\bar{x}, \bar{y}, \bar{z})$ is the centroid of all points in $\neifunc_{\voxnet}(\bx)$.
Then aggregation (Equation \ref{equ:voxelnet_2}) is done by 
a pointwise fully connected layer $\fcn^{(\kthagglayer)}(\cdot)$ 
followed by an element-wise max pooling 
$\maxpool_{\bx_i \in \neifunc_{\voxnet}(\bx)}\{\cdot\}$. 
$\aggemb^{(\kthagglayer)}_{\bx}$ encodes the shape contained within the current voxel $\neifunc_{\voxnet}(\bx)$. 
$\comfunc(\cdot, \cdot)$ is simply a vector concatenation operation (See Equation \ref{equ:voxelnet_3}), 
and $\readoutfunc_{\voxnet}(\cdot)$ is a identity function (See Equation \ref{equ:voxelnet_4}).

\textbf{\sagatmodel}.
\citet{mai2020multi} proposed a modified graph attention network (GAT) \citep{velivckovic2018graph} to model the spatial interactions among nearby locations (e.g., POIs). 
$\neifunc_{\knn}(\bx)$ is defined as all k nearest neighbors of location $\bx$.
The original model focuses on encoding feature information $\bv$ for each location such as POI type, 
but here we generalize it as a generic local aggregation location encoder
(Spatial-Aware Graph Attention Network, in short, \textbf{\sagatmodel}):
	\begin{subequations}
		\label{equ:sagat}
		\begin{align}
			\locemb^{(0)}_{\bx} & = \enc_{\sagat}(\bx), \label{equ:sagat_1} \\	
			\aggemb^{(\kthagglayer)}_{\bx} &= \aggfunc^{(\sagat)}_{\bx_i \in \neifunc_{\knn}(\bx)}\{\locemb^{(\kthagglayer-1)}_{\bx_i}\}
			= \attnact(\dfrac{1}{\attnsize} \sum_{\attnidx=1}^{\attnsize} \sum_{\bx_i \in \neifunc_{\knn}(\bx)} \attnscr^{(\kthagglayer)}_{i \attnidx} \locemb^{(\kthagglayer-1)}_{\bx_i}) , \label{equ:sagat_2} \\
			\locemb^{(\kthagglayer)}_{\bx} & = \comfunc^{(\sagat)}(\locemb^{(\kthagglayer-1)}_{\bx}, \aggemb^{(\kthagglayer)}_{\bx}) = \aggemb^{(\kthagglayer)}_{\bx},  \label{equ:sagat_3} \\
			\enc^{(\mP)}_{\sagat}(\bx)   & =  \readoutfunc^{(\sagat)}(\locemb^{(\numagglayer)}_{\bx}) = \locemb^{(\numagglayer)}_{\bx} , \label{equ:sagat_4} \\
			where \; \attnscr^{(\kthagglayer)}_{i \attnidx} & =  \frac{exp(\sigma^{(\kthagglayer)}_{i \attnidx})}{\sum_{\bx_o \in \neifunc_{\knn}(\bx) } exp(\sigma^{(\kthagglayer)}_{o\attnidx})} , \label{equ:sagat_5} \\
			\sigma^{(\kthagglayer)}_{i \attnidx} & = \leakyrelu( (\attnvec^{(\kthagglayer)}_{\attnidx})^{T}[\locemb^{(\kthagglayer-1)}_{\bx}; \locemb^{(\kthagglayer-1)}_{\bx_i};\enc(\bx - \bx_{i}) ]). \label{equ:sagat_6} 		
		\end{align}
	\end{subequations}
In Equation \ref{equ:sagat_1}, \sagatmodel~ first uses $\enc_{\sagat}(\bx)$ to lift each location into the embedding space. 
Originally, \citet{mai2020multi} used a feature encoder as $\enc_{\sagat}(\bx)$ to represent $\bv$ into a feature embedding. However, here, we define $\enc_{\sagat}(\bx)$ to be any single point location encoder discussed in Section \ref{sec:single-pt}. 
Equation \ref{equ:sagat_2} is the most important neighborhood aggregation step. 
Multi-head attention is adopted to aggregate the neighboring location embedding $ \locemb^{(\kthagglayer-1)}_{\bx_i}$ of $\bx_i \in \neifunc_{\knn}(\bx)$ from the previous layer. $\attnsize$ is the total number of attention heads. $\attnscr^{(\kthagglayer)}_{i \attnidx}$ is the attention coefficient of $ \locemb^{(\kthagglayer-1)}_{\bx_i}$ in the $\attnidx$th attention head within the $\kthagglayer$th layer which is normalized across $\neifunc_{\knn}(\bx)$ based on Equation \ref{equ:sagat_5}. $\sigma^{(\kthagglayer)}_{i \attnidx}$ is its non-normalized counterpart.
In Equation \ref{equ:sagat_6}, $\locemb^{(\kthagglayer-1)}_{\bx}$ and $\locemb^{(\kthagglayer-1)}_{\bx_i}$ are the representations of the center location $\bx$ and its neighbor $\bx_i$ in the $\kthagglayer-1$-th layer.
Compared with GAT \citep{velivckovic2018graph},
$\enc(\bx - \bx_{i})$ is added in the attention score computation so the spatial affinity $\Delta \bx_i = \bx - \bx_{i}$ is considered in the aggregation step.  This practice makes \sagatmodel~ ``spatial-aware''. $\attnvec^{(\kthagglayer)}_{\attnidx}$ is a learnable attention vector for the $\attnidx$th attention head and $\leakyrelu(\cdot)$ is the LeakyReLU activation function. $\enc(\cdot)$ is an encoder for spatial relation,
which can be any single-point location encoder such as $\tile$, $\rbf$, $\direct$, $\theory$, $\grid$, $\protein$, and so on. 
\citet{mai2020multi} only utilized  one aggregation layer, i.e., $\numagglayer = 1$. Here, we generalize it into multiple layers, i.e., $\numagglayer \geq 1$. The combination and readout function are both identity functions.

\begin{figure*}[t!]
	\centering 
	\includegraphics[width=0.6\textwidth]{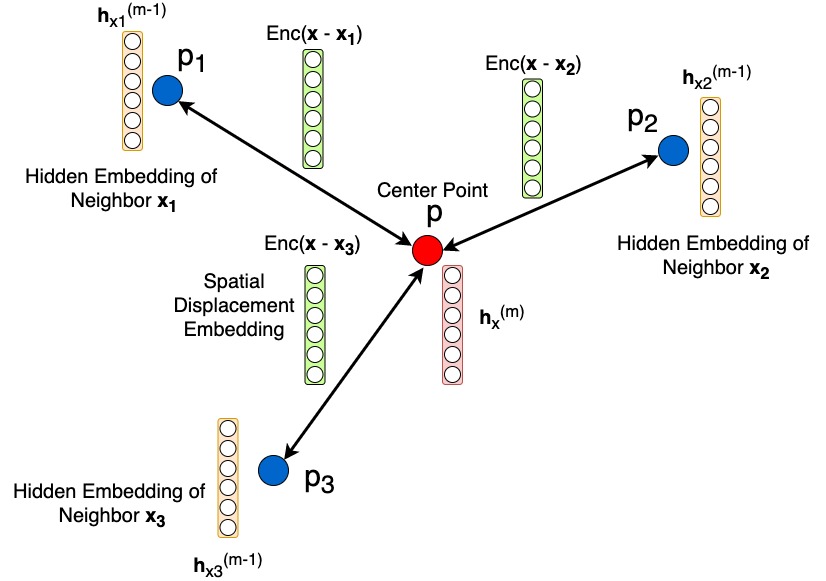}
	\caption[]%
	{
		An illustration of the $\kthagglayer$th aggregation layer of \sagatmodel.
	}
	\label{fig:sagat}
\end{figure*}

Figure \ref{fig:sagat} illustrates the $\kthagglayer$th aggregation layer of \sagatmodel. In this example, we consider three neighbors. Three yellow vectors $\locemb^{(\kthagglayer-1)}_{\bx_i}$ ($i=1,2,3$) are the hidden embeddings of three neighbors. Three green vectors $\enc(\bx - \bx_{i})$ are the spatial displacement embeddings which are used in the spatial-aware graph attention attention. The red vector are the hidden embedding of center location $\bx$ in the next layer.

\textbf{\dgcnnmodel}.
Dynamic Graph CNNs (in short, \dgcnnmodel) \citep{wang2019dynamic} is proposed to jointly consider global shape structure and local neighborhood information when learning on point clouds. \dgcnnmodel~ also has $\numagglayer$ LEA layers.
Compared with other location encoders, \dgcnnmodel~ has two crucial distinctions: 1) dynamic graph neighborhood $\neifunc^{(\kthagglayer)}_{\dgcnn}(\bx)$ and 2) an edge convolution module $\edgeconv^{(\kthagglayer)}_{\bx_i \in \neifunc^{(\kthagglayer)}_{\dgcnn}(\bx)}\{\cdot\}$.

	\begin{subequations}
		\label{equ:dgcnn}
		\begin{align}
			\aggemb^{(\kthagglayer)}_{\bx} &= \aggfunc^{(\dgcnn)}_{\bx_i \in \neifunc^{(\kthagglayer)}_{\dgcnn}(\bx)}\{\locemb^{(\kthagglayer-1)}_{\bx_i}\} = \edgeconv^{(\kthagglayer)}_{\bx_i \in \neifunc^{(\kthagglayer)}_{\dgcnn}(\bx)}\{\locemb^{(\kthagglayer-1)}_{\bx_i}\}
			 \label{equ:dgcnn_2} \\
			& = \maxpool_{\bx_i \in \neifunc^{(\kthagglayer)}_{\dgcnn}(\bx)}\{\mlp^{(\kthagglayer)}_{e}([\locemb^{(\kthagglayer-1)}_{\bx} ; \locemb^{(\kthagglayer-1)}_{\bx_i} - \locemb^{(\kthagglayer-1)}_{\bx}])\}, \label{equ:dgcnn_3} \\
			\locemb^{(\kthagglayer)}_{\bx} & = \comfunc^{(\dgcnn)}(\locemb^{(\kthagglayer-1)}_{\bx}, \aggemb^{(\kthagglayer)}_{\bx}) = \aggemb^{(\kthagglayer)}_{\bx}.  \label{equ:dgcnn_4} 
		\end{align}
	\end{subequations}
The former means that instead of using a fixed neighborhood as VoxelNet and \sagatmodel~ does, \dgcnnmodel~ recomputes the neighborhood of $\bx$ for each LEA layer. In the $\kthagglayer$th layer, given the embeddings of all locations from the previous layer $\locembset^{(\kthagglayer-1)} = \{\locemb^{(\kthagglayer-1)}_{\bx_j} | \forall \bx_j \in \mP\}$, $\neifunc^{(\kthagglayer)}_{\dgcnn}(\bx)$ is defined as the top K nearest neighbors of $\locemb^{(\kthagglayer-1)}_{\bx}$ from $\locembset^{(\kthagglayer-1)}$ in the embedding space (not Euclidean space). Since $\locembset^{(\kthagglayer-1)}$ is updated after each LEA layer, $\neifunc^{(\kthagglayer)}_{\dgcnn}(\bx)$ is called a dynamic neighborhood. 
As for the edge convolution, $\edgeconv^{(\kthagglayer)}_{\bx_i \in \neifunc^{(\kthagglayer)}_{\dgcnn}(\bx)}\{\cdot\}$ is an aggregation operator which captures local geometric structure while maintaining permutation invariance. Equation \ref{equ:dgcnn} mainly shows how the edge convolution works while skipping other steps such as location embedding initialization and the readout function. $\edgeconv^{(\kthagglayer)}_{\bx_i \in \neifunc^{(\kthagglayer)}_{\dgcnn}(\bx)}\{\cdot\}$ concatenates $\locemb^{(\kthagglayer-1)}_{\bx}$ and its affinity to its neighbor $\locemb^{(\kthagglayer-1)}_{\bx_i} - \locemb^{(\kthagglayer-1)}_{\bx}$. 
The result is fed into a multi-layer perceptron $\mlp^{(\kthagglayer)}_{e}(\cdot)$ followed by a pointwise max pooling. 
The combination operator is an identity function (See Equation \ref{equ:dgcnn_4}).
Although $\neifunc^{(\kthagglayer)}_{\dgcnn}(\bx)$ is dynamic, we only consider single-scale neighborhoods for each $\bx$. So we classify \dgcnnmodel~ can a local aggregation location encoder.

\begin{figure*}[ht!]
	\centering 
	\includegraphics[width=0.5\textwidth]{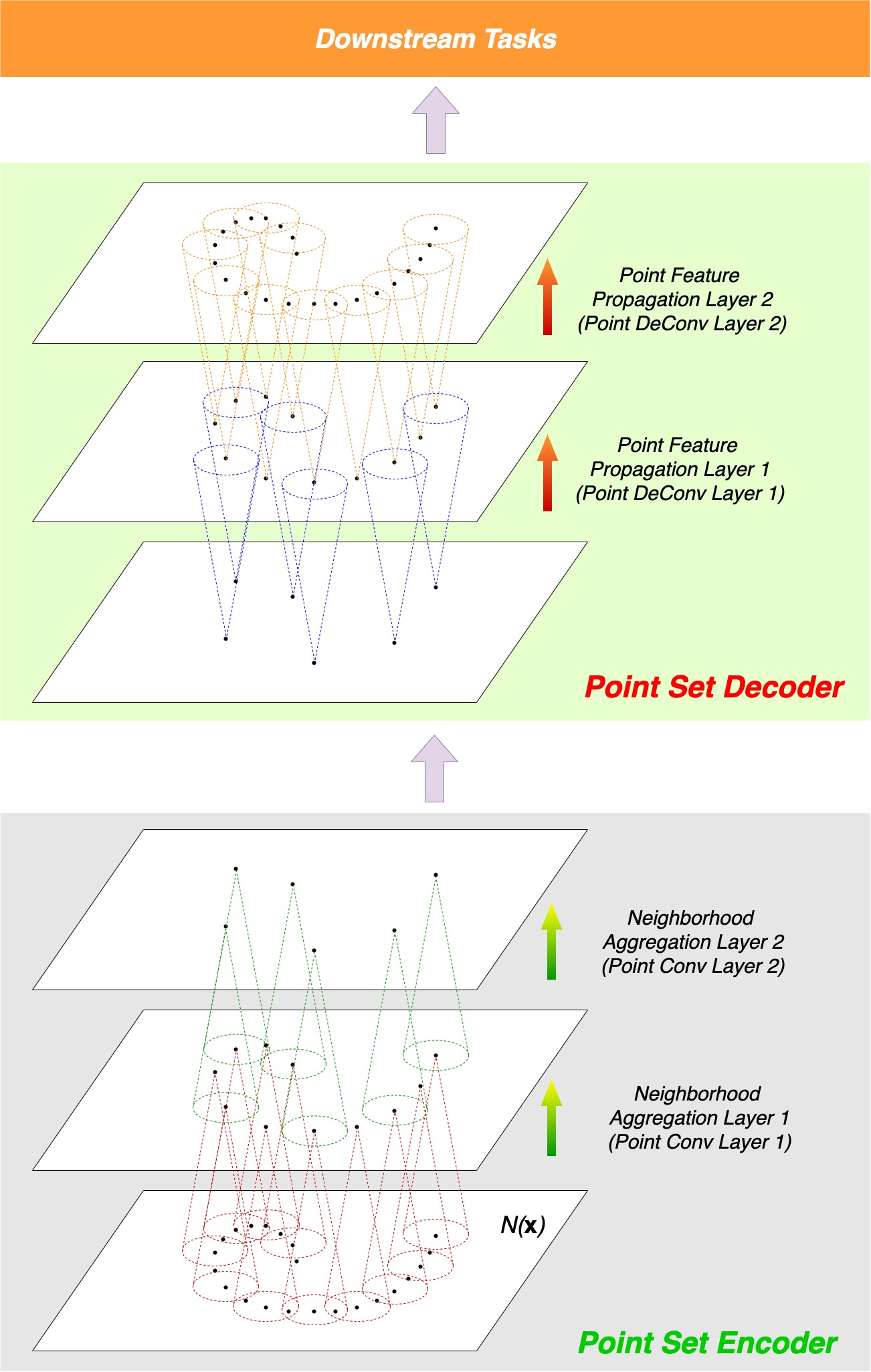}
	\caption[]%
	{
 An illustration of the Conv-DeConv architecture \citep{noh2015learning}  for $\enc^{(\mP)}_{\multiagg}(\bx)$ such as PointNet++ \citep{qi2017pointnet++}, PointCNN \citep{li2018pointcnn}, and Graph-Conv GAN \citep{valsesia2018learning}. 
	}
	\label{fig:multi_agg_loc_enc}
\end{figure*}

\subsubsection{Hierachical aggregation location encoder}   \label{sec:multi-scale-agg}

Instead of defining one neighborhood per location, the hierachical 
aggregation location encoder $\enc^{(\mP)}_{\multiagg}(\bx)$ defines a multi-scale neighborhood for location $\bx$, i.e., $\neifunc_{\multiagg}(\bx) =   \{ \neifunc^{(1)}_{\multiagg}(\bx), \neifunc^{(2)}_{\multiagg}(\bx),..., \neifunc^{(\kthagglayer)}_{\multiagg}(\bx),..., \neifunc^{(\numagglayer)}_{\multiagg}(\bx) \}$ which can be aggregated in a hierarchical manner as illustrated in Figure \ref{fig:multi_agg_loc_enc}.

\begin{definition}[Hierachical Aggregation Location Encoder]
   \label{def:multiagg-locenc}
   
   $\enc^{(\mP)}_{\multiagg}(\bx)$ first hierarchically aggregates location 
   features from these neighborhoods and then propagates the aggregated features back to each location.
   $\enc^{(\mP)}_{\multiagg}(\bx)$ usually adopts a Conv-DeConv \citep{noh2015learning} like architecture which consists of two modules:
    1) \textbf{Point Set Encoder $\ptconv^{(\mP)}_{\multiagg}(\cdot)$}: a stack of $\numagglayer$ neighborhood aggregation layers (or called Point Conv layers \citep{li2018pointcnn}) in which each location embedding in the $\kthagglayer$ layer is aggregated from its neighborhood $\neifunc^{(\kthagglayer-1)}_{\multiagg}(\bx)$ in the previous layer; 2)  \textbf{Point Set Decoder $\ptdeconv^{(\mP)}_{\multiagg}(\cdot)$}: a stack of $\numagglayer$ point feature propagation layers (or called Point DeConv layers) in which each DeConv-like architecture propagates the location embeddings from the previous layers to a denser set of locations. This whole architecture follows the U-Net design \citep{ronneberger2015u}: 
   
   \begin{align}
		\enc^{(\mP)}_{\multiagg}(\bx) = \ptdeconv^{(\mP)}_{\multiagg}(\ptconv^{(\mP)}_{\multiagg}(\bx))
	\label{equ:multiagg}
	\end{align}
	Here, $\ptconv^{(\mP)}_{\multiagg}(\cdot)$ follows the model setup in Equation \ref{equ:aggenc}. 
	Sometimes, $\ptdeconv^{(\mP)}_{\multiagg}(\cdot)$ also follows Equation \ref{equ:aggenc} such as PointCNN \citep{li2018pointcnn}.
\end{definition}

Figure \ref{fig:multi_agg_loc_enc} shows an illustration of this Conv-DeConv architecture of $\enc^{(\mP)}_{\multiagg}(\bx)$.

\textbf{PointNet++}. 
As an extension of PointNet, 
PointNet++ \citep{qi2017pointnet++} follows the Conv-DeConv like architecture in Figure \ref{fig:multi_agg_loc_enc} to achieve hierarchical feature aggregation and propagation.
The point set encoder $\ptconv^{(\mP)}_{\poinetplus}(\bx)$ consists of
$\numagglayer$ Point Conv Layers (so-called set abstraction levels (SAL) in \citet{qi2017pointnet++}) each of which is composed of three key layers:

\begin{enumerate}
	\item \textbf{The sampling layer} at the $\kthagglayer$th SAL samples a point subset $\mP^{(\kthagglayer)}$ from 
	the previous SAL -  $\mP^{(\kthagglayer-1)}$, and, therefore, $\mP^{(\kthagglayer)} \subset \mP^{(\kthagglayer-1)}$ and $\mP^{(0)} = \mP$. 
	
	\item \textbf{The grouping layer} at the $\kthagglayer$th SAL groups points in $\mP^{(\kthagglayer-1)}$ into 
	the neighborhood $\neifunc_{\poinetplus}^{(\kthagglayer)}(\bx) \subseteq \mP^{(\kthagglayer-1)}$ of each point $\bx \in \mP^{(\kthagglayer)}$. 
	$\neifunc_{\poinetplus}^{(\kthagglayer)}(\bx)$ is defined as all points $\bx_i \in \mP^{(\kthagglayer-1)}$ within 
	radius $\buffradius^{(\kthagglayer)}$.
	
	\item \textbf{The aggregation layer} at the $\kthagglayer$th SAL produces new location embedding $\locemb^{(\kthagglayer)}_{\bx}$ for each $\bx \in \mP^{(\kthagglayer)}$ by aggregating  $\locemb^{(\kthagglayer-1)}_{\bx_i}$ of $\bx_i$ in $\bx$'s neighborhood $\neifunc_{\poinetplus}^{(\kthagglayer)}(\bx)$. 
\end{enumerate}

\begin{subequations}
    \vspace{-0.5cm}
	\label{equ:ssg}
	\begin{align}
	\locemb^{(0)}_{\bx} & = \enc_{\direct}(\bx), \label{equ:ssg_1}\\
	\aggemb^{(\kthagglayer)}_{\bx} & = \aggfunc^{(\ssg)}_{\bx_i \in \neifunc_{\poinetplus}^{(\kthagglayer)}(\bx)}\{ \locemb^{(\kthagglayer-1)}_{\bx_i} \} \\
	& =  \maxpool_{\bx_i \in \neifunc_{\poinetplus}^{(\kthagglayer)}(\bx)}\{\mlp^{(\kthagglayer)}( \locemb^{(\kthagglayer-1)}_{\bx_i}  )\},  \label{equ:ssg_2} \\
	\locemb^{(\kthagglayer)}_{\bx} & = \comfunc^{(\ssg)}(\locemb^{(\kthagglayer-1)}_{\bx}, \aggemb^{(\kthagglayer)}_{\bx} ) = \aggemb^{(\kthagglayer)}_{\bx}, \label{equ:ssg_3} \\
	\ptconv^{(\mP)}_{\ssg}(\bx) &= \readoutfunc^{(\ssg)}(\locemb^{(\numagglayer)}_{\bx}) = \locemb^{(\numagglayer)}_{\bx}. \label{equ:ssg_4}
	\end{align}
\end{subequations}

\citet{qi2017pointnet++} proposed several versions of PointNet++: \textbf{SSG} (ablated PointNet++ with single scale grouping in each level), \textbf{MSG} (multi-scale grouping PointNet++), and \textbf{MRG} (multi resolution grouping PointNet++). All of them are $\enc^{(\mP)}_{\multiagg}(\bx)$ according to Definition \ref{def:multiagg-locenc}.
Equation \ref{equ:ssg} shows how the point set encoder of \textbf{SSG} works. In the $\kthagglayer$th SAL, the PointNet layer 
aggregates features in $\neifunc_{\poinetplus}^{(\kthagglayer)}(\bx)$ by using a PointNet-like aggregation function $\aggfunc^{(\ssg)}_{\bx_i \in \neifunc_{\poinetplus}^{(\kthagglayer)}(\bx)}\{\cdot\}$ - $\mlp^{(\kthagglayer)}(\cdot)$ followed by a max pooling function (See Equation \ref{equ:ssg_2}).

Because $\mP^{(\numagglayer)} \subset \mP^{(\numagglayer-1)} \subset ... \subset \mP^{(0)} = \mP$, $\mP^{(\numagglayer)}$ is a small subset of $\mP$ which can be seen as the skeleton points of $\mP$.
$\ptconv^{(\mP)}_{\ssg}(\cdot)$ can only produce embeddings for 
$\bx \in \mP^{(\numagglayer)}$. To obtain embeddings for each $\bx \in \mP$, 
$\ptdeconv^{(\mP)}_{\ssg}(\cdot)$ is used to propagate location features back to each $\bx \in \mP$ by using an inverse distance weighted interpolation method. This can be seen as a reverse process of $\ptconv^{(\mP)}_{\ssg}(\cdot)$. 
The sampled point sets $\{ \mP^{(0)} = \mP, \mP^{(1)}, \mP^{(2)},...,\mP^{(\kthagglayer)},...,\mP^{(\numagglayer)}   \}$ 
are used in a reverse manner to progressively interpolate the location features back to a larger sampled point set until we get location 
embeddings for all $\bx \in \mP$. This idea follows the Conv-DeConv 
idea in Equation \ref{equ:multiagg}.

Compared with \textbf{SSG}, \textbf{MSG} changes the point set encoder by concatenating location embeddings of the same points obtained from neighborhoods with different spatial scales. \textbf{MSG} is rather computationally expensive since it aggregates features in larger scale neighborhoods for each centroid point. \textbf{MRG} solves this by concatenating location embeddings obtained from different \textbf{SSG} point set encoders with varied numbers of SAL layers. Please refer to \citet{qi2017pointnet++} for detailed description.

\textbf{PointCNN}. Similar to PointNet++, PointCNN \citep{li2018pointcnn} also utilizes a point set encoder with $\numagglayer$ Point Conv layers
(which 
we call PointCNN Layers here). Each layer also consists of three key steps: sampling layer, grouping layer, aggregation layer (PointNet layer). The differences from \textbf{SSG} 
are mainly in grouping and aggregation layer. 
The $\kthagglayer$th PointCNN grouping layer 
defines the neighborhood $\neifunc^{(\kthagglayer)}_{\pointcnn}(\bx) \subset \mP^{(\kthagglayer-1)}$ as a set of $\topk$ points uniformly sampled from the $\topk \dilate$ nearest neighbors of $\bx$ obtained from $ \mP^{(\kthagglayer-1)}$.
This can be seen as an analogy of the dilated convolution idea from the traditional CNN models and $\dilate$ is the dilation rate.
The PointCNN aggregation layer defines a convolution operation, $\conv^{(\pointcnn)}(\convkernel^{(\kthagglayer)}, \cdot)$,  over $\neifunc^{(\kthagglayer)}_{\pointcnn}(\bx)$ 
as an analogy to the convolution operation over images. 
$\convkernel^{(\kthagglayer)}$ is the convolution kernels in the $\kthagglayer$th  
layer.

\begin{subequations}
    \vspace{-0.5cm}
	\label{equ:pointcnn}
	\begin{align}
	\locemb^{(\kthagglayer)}_{\bx_i,\delta} & = \mlp^{(\kthagglayer)}_{\delta}(\bx_i - \bx), \forall \bx_i \in \neifunc^{(\kthagglayer)}_{\pointcnn}(\bx), 
	\label{equ:pointcnn_mlp1}\\
	\xtrans^{(\kthagglayer)} & = \mlp^{(\kthagglayer)}(\rowconcat_{\neifunc^{(\kthagglayer)}_{\pointcnn}(\bx)}(\bx_i - \bx)), 
	\label{equ:pointcnn_mlp2}\\
	\aggemb^{(\kthagglayer)}_{\bx} & = \aggfunc^{(\pointcnn)}_{\bx_i \in \neifunc^{(\kthagglayer)}_{\pointcnn}(\bx)}\{\locemb^{(\kthagglayer-1)}_{\bx_i}\} \\
	& = \conv^{(\pointcnn)}(\convkernel^{(\kthagglayer)}, \xtrans^{(\kthagglayer)} \times \rowconcat_{\neifunc^{(\kthagglayer)}_{\pointcnn}(\bx)}([\locemb^{(\kthagglayer)}_{\bx_i,\delta}; \locemb^{(\kthagglayer-1)}_{\bx_i}]))
	, \label{equ:pointcnn_conv} \\
	\locemb^{(\kthagglayer)}_{\bx} & = \comfunc^{(\pointcnn)}(\locemb^{(\kthagglayer-1)}_{\bx}, \aggemb^{(\kthagglayer)}_{\bx}) =  \aggemb^{(\kthagglayer)}_{\bx}, \label{equ:pointcnn_cmb} \\
	\ptconv^{(\mP)}_{\pointcnn}(\bx) &= \readoutfunc^{(\pointcnn)}(\locemb^{(\numagglayer)}_{\bx}) = [\locemb^{(\numagglayer)}_{\bx}; \mlp_{g}(\bx)], \label{equ:pointcnn_rdt}
	\end{align}
\end{subequations}

Equation \ref{equ:pointcnn} describes how the point set encoder of PointCNN works.
In Equation \ref{equ:pointcnn_mlp1}, a multiple-layer perceptron $\mlp^{(\kthagglayer)}_{\delta}(\cdot)$ individually lifts the spatial affinity $\bx_i - \bx$ for each neighbor $\bx_i \in \neifunc^{(\kthagglayer)}_{\pointcnn}(\bx)$ into $\locemb^{(\kthagglayer)}_{\bx_i,\delta} \in \Real^{C_\delta}$, a $C_{\delta}$ dimensional embedding. 
Then in Equation \ref{equ:pointcnn_mlp2}, $\rowconcat_{\neifunc^{(\kthagglayer)}_{\pointcnn}(\bx)}(\bx_i - \bx) \in \Real^{\topk \times \spdim}$ represents a stack of the spatial affinity vector $\bx_i - \bx \in \Real^{\spdim}$ of all 
 $\bx_i \in \neifunc^{(\kthagglayer)}_{\pointcnn}(\bx)$ which results in a $\topk \times \spdim$ matrix. The multi-layer perceptron $\mlp^{(\kthagglayer)}(\cdot)$ 
converts this matrix into a $\topk \times \topk$ $\xtrans$-transformation matrix - $\xtrans^{(\kthagglayer)}$. 
Next, in Equation \ref{equ:pointcnn_conv}, 
a point convolution operator $\conv^{(\pointcnn)}(\convkernel^{(\kthagglayer)}, \cdot)$
aggregates the concatenation $[\locemb^{(\kthagglayer)}_{\bx_i,\delta}; \locemb^{(\kthagglayer-1)}_{\bx_i}] \in \Real^{C_{\delta} + \embdim^{(\kthagglayer-1)}}$. 
$\xtrans^{(\kthagglayer)}$ is used here to permute this $\topk \times (C_{\delta} + \embdim^{(\kthagglayer-1)})$ 
matrix and has to be aware of the order of all $\bx_i \in \neifunc^{(\kthagglayer)}_{\pointcnn}(\bx)$.
In the final readout function (Equation \ref{equ:pointcnn_rdt}) another 
$\mlp_{g}(\cdot)$ is used to directly lift $\bx$ into a high dimensional embedding which is concatenated with $\locemb^{(\numagglayer)}_{\bx} \in \Real^{\embdim^{(\numagglayer)}}$. This can be seen as an analogy of the skip-connection in traditional CNN. 
In Equation \ref{equ:pointcnn_conv}, when $\kthagglayer=1$, $\locemb^{(\kthagglayer-1)}_{\bx_i} = \locemb^{(0)}_{\bx_i}$ needs to be initialized. Given $\pt_i = (\bx_i, \bv_i)$ 
(See Definition \ref{def:locenc}), 
\citet{li2018pointcnn} made $\locemb^{(0)}_{\bx_i}= \bv_i$. We also can choose $\locemb^{(0)}_{\bx_i}= \bx_i$.

Similar PointCNN layers are deployed in the point set decoder $\ptdeconv^{(\mP)}_{\pointcnn}(\bx)$ to form a Conv-DeConv like architecture. Similar $\conv^{(\pointcnn)}(\convkernel^{(\kthagglayer)}, \cdot)$ operator is used while the only difference is $\ptdeconv^{(\mP)}_{\pointcnn}(\bx)$ has more points but
less feature channels in its output vs. its input, and $\{\mP^{(\kthagglayer)}\}$ are forwarded from $\ptconv^{(\mP)}_{\pointcnn}(\bx)$.

\textbf{Graph-Conv GAN}. 
\citet{valsesia2018learning} presented a graph convolution based $\enc^{(\mP)}_{\multiagg}(\bx)$ 
for 3D point cloud generation based on Generative Adversarial Network (GAN) \citep{goodfellow2014generative}.
We denote the location encoder of the model as $\enc^{(\mP)}_{\gcgan}(\bx)$.
The same Conv-DeConv 
idea is used here (See Definition \ref{def:multiagg-locenc}) and 
$\ptconv^{(\mP)}_{\gcgan}(\bx)$
consists of $\numagglayer$ Point Conv layers each of which is composed of a sampling, a grouping, and an aggregation layer: 
\small
\begin{subequations}
	\label{equ:gc-gan-agg}
	\begin{align}
	\locemb^{(0)}_{\bx} & = \enc(\bx), \label{equ:gc-gan_1}\\
	\aggemb^{(\kthagglayer)}_{\bx} & = \aggfunc^{(\gcgan)}_{\bx_i \in \neifunc^{(\kthagglayer)}_{\knn}(\bx)}\{\locemb^{(\kthagglayer-1)}_{\bx_i}\} = \sum_{\bx_i \in \neifunc^{(\kthagglayer)}_{\knn}(\bx)}{ \dfrac{ F^{(\kthagglayer)} (\locemb^{(\kthagglayer-1)}_{\bx_i} - \locemb^{(\kthagglayer-1)}_{\bx})  \locemb^{(\kthagglayer-1)}_{\bx_i} }{| \neifunc^{(\kthagglayer)}_{\knn}(\bx) |} }
	, \label{equ:gc-gan_2} \\
	\locemb^{(\kthagglayer)}_{\bx} & = \comfunc^{(\gcgan)}(\locemb^{(\kthagglayer-1)}_{\bx}, \aggemb^{(\kthagglayer)}_{\bx}) =  \sigma \Big( \aggemb^{(\kthagglayer)}_{\bx} +   \mathbf{W}^{(\kthagglayer)} \locemb^{(\kthagglayer-1)}_{\bx} + \mathbf{b}^{(\kthagglayer)} \Big), \label{equ:gc-gan_3}\\
	\ptconv^{(\mP)}_{\gcgan}(\bx)&  = \readoutfunc^{(\gcgan)}(\locemb^{(\numagglayer)}_{\bx}) = \locemb^{(\numagglayer)}_{\bx}.
	\label{equ:gc-gan_4} 	
	\end{align}
\end{subequations} \normalsize
We denote the location embedding of $\bx$ at the $\kthagglayer$th layer as $\locemb^{(\kthagglayer)}_{\bx} \in \Real^{\embdim^{(\kthagglayer)}}$. 
$\enc(\bx)$ in Equation \ref{equ:gc-gan_1} can be any single point location encoder. 
$\aggfunc^{(\gcgan)}_{\bx_i \in \neifunc^{(\kthagglayer)}_{\knn}(\bx)}\{\locemb^{(\kthagglayer-1)}_{\bx_i}\} \in \Real^{\embdim^{(\kthagglayer)}}$ in Equation \ref{equ:gc-gan_2} uses a graph convolution operator to aggregate the neighborhood $\neifunc^{(\kthagglayer)}_{\knn}(\bx)$ at the $\kthagglayer$th layer. 
$\neifunc^{(\kthagglayer)}_{\knn}(\bx)$ is defined as $\bx$'s $\topk$-th nearest neighbors - $\neifunc^{(\kthagglayer)}_{\knn}(\bx) = \KNNfunc(\bx, \topk, \mP^{(\kthagglayer-1)})$. 
$\locemb^{(\kthagglayer-1)}_{\bx_i} - \locemb^{(\kthagglayer-1)}_{\bx} \in \Real^{\embdim^{(\kthagglayer-1)}}$ capture the spatial affinity between neighboring location $\bx_i$ and $\bx$. $F^{(\kthagglayer)} (\cdot)$ denotes a fully-connected network 
which 
regresses $\locemb^{(\kthagglayer-1)}_{\bx_i} - \locemb^{(\kthagglayer-1)}_{\bx}$ into
a matrix $F^{(\kthagglayer)} (\locemb^{(\kthagglayer-1)}_{\bx_i} - \locemb^{(\kthagglayer-1)}_{\bx}) \in \Real^{\embdim^{(\kthagglayer)} \times \embdim^{(\kthagglayer-1)}}$.
Equation \ref{equ:gc-gan_3} shows how to combine the neighborhood feature $\aggemb^{(\kthagglayer)}_{\bx} \in \Real^{\embdim^{(\kthagglayer)}}$ with $\bx$'s own feature from the previous layer $\locemb^{(\kthagglayer-1)}_{\bx} \in \Real^{\embdim^{(\kthagglayer-1)}}$. $\mathbf{W}^{(\kthagglayer)} \in \Real^{\embdim^{(\kthagglayer)} \times \embdim^{(\kthagglayer-1)}}$ and $\mathbf{b}^{(\kthagglayer)} \in \Real^{\embdim^{(\kthagglayer)}}$ are a learnable matrix and a bias vector respectively.

It is worth mentioning that the main difference among SSG, PointCNN, and Graph-Conv GAN is different aggregation layers used in their point set encoders (See Equation \ref{equ:ssg}, \ref{equ:pointcnn}, and \ref{equ:gc-gan-agg}). Figure \ref{fig:multi_agg_loc_enc} shows their shared Conv-DeConv architecture.

\subsection{Comparison among different models}   \label{sec:compare}

After introducing $\enc(\bx)$ and $\enc^{(\mP)}(\bx)$, it is worth to compare them from different aspects. First, we provide a general comparison between $\enc(\bx)$ and $\enc^{(\mP)}(\bx)$:
\begin{enumerate}
\item 
$\enc(\bx)$ 
encodes $\bx$ independently without considering its spatial context. In contrast, $\enc^{(\mP)}(\bx)$ jointly consider $\bx$ and its neighbor $\neifunc(\bx)$. $\enc^{(\mP)}(\bx)$ can be seen as a generalizaton of $\enc(\bx)$ in which any $\enc(\bx)$ can be used to compute $\locemb^{(0)}_{\bx}$ (See Equation \ref{equ:aggenc}).
\item 
When a new location $\bx^{\prime}$ is added to $\mP$, $\enc(\bx_i)$ is unaffected for all $\bx_i \in \mP$. In contrast, as for $\enc^{(\mP)}(\bx)$, for all $\bx_i \in \{\bx_i | \bx_i \in \mP \wedge \bx^{\prime} \in \neifunc(\bx_i) \}$, $\enc^{(\mP)}(\bx_i)$ will be updated since $\neifunc(\bx_i)$ is modified. However, $\enc^{(\mP)}_{\kernel}(\bx)$ is unaffected since $\neifunc(\bx) = \mQ $ is the same for all $\bx \in \mP$ which is unchanged. 

\item 
$\enc(\bx)$ has a rather high inference speed, while the aggregation operator in $\enc^{(\mP)}(\bx)$ is time-consuming.

\item 
Since $\enc^{(\mP)}(\bx)$ additionally considers $\neifunc(\bx)$, it has richer features for model prediction and has a potential higher performance compared to $\enc(\bx)$.
\end{enumerate}

We can see that both $\enc(\bx)$ and $\enc^{(\mP)}(\bx)$ have advantages and disadvantages. Although both of them are task-agnostic, they excel at different tasks and the model selection should depend on the current task: 
\begin{enumerate}
	\item The first criterion 
	is the input for a single model prediction - one location $\bx$ (e.g., geo-aware image classification) or a whole point set $\mP$ (e.g., point cloud segmentation). The former setup prefers $\enc(\bx)$ given its fast inference speed. The latter one prefers $\enc^{(\mP)}(\bx)$ since it can additionally captures spatial context information. Moreover, some tasks require producing a single embedding for the whole point set $\mP$ or parts of it such as point cloud classification and objection recognition. For these tasks, $\enc^{(\mP)}(\bx)$ is the only choice.
	
	\item Another criterion is the preference between faster inference speed or higher prediction accuracy. 
	A mobile application may prefer a faster inference speed in which $\enc(\bx)$ excels. An enterprise application might prefer higher prediction accuracy where $\enc^{(\mP)}(\bx)$ is preferred.
\end{enumerate}

Next, we compare different sub-categories of location encoders from five different perspectives. The results shown in Table \ref{tab:review} will be discussed in detail as follows.
\begin{enumerate}
	\item $\spdim$: In terms of the spatial dimension of $\bx$ a location encoder can handle, almost all models can handle different $\spdim$, e.g., $\spdim = 2, 3$. Exceptions are GPS2Vec, $\aodha$, and $\theory$. GPS2Vec and $\aodha$ are specifically designed for GPS coordinates which can be uniquely identified by $\lat$ and $\lon$. 
	$\theory$ is designed only for 2D coordinates since it is inspired by neuroscience research about grid cells which are critical for self-motion integration and navigation of mammals in a 2D space.
	\item Parametric: As shown in Table \ref{tab:review}, all models except $\rbf$ and adaptive kernel$^*$ are parametric models, which means their learnable parameters have a fixed size. $\rbf$ and adaptive kernel$^*$ can be either parametric or non-parametric models. Since $\locemb^{(1)}_{\bx} \in \Real^{| \mQ |}$ (See Equation \ref{equ:kernel_2}, \ref{equ:kernel_3}), the size of the kernel center set $\mQ$ decides the number of learnable parameters in $\pemlp(\cdot)$ of $\enc^{(\mP)}_{\kernel}(\bx)$. If $\mQ = \mP$, then both $\rbf$ and adaptive kernel$^*$ become non-parametric models. Otherwise, if $\mQ$ has a fixed size, both of them are parametric model.

	\item Multi-scale: 
	Both $\enc_{\sinmulti}(\bx)$ and $\enc^{(\mP)}_{\multiagg}(\bx)$ utilize multi-scale approaches. So they are better at capturing locations with non-uniform density or a mixture of distributions with different characteristics. However, they adopt different multi-scale approaches. The former designs multi-scale representations based on $\psenc(\bx)$ which uses sinusoidal functions with different frequencies, while the latter aggregates the neighborhood of $\bx$ in a hierarchical manner. From a practical perspective, many studies have shown that multi-scale location encoders can outperform single-scale models and models without scale related parameters on various tasks. For example, \citet{qi2017pointnet++} showed that PointNet++ can outperform PointNet on both point cloud classification and segmentation task. \citet{mai2020multi} showed that multi-scale models ($\theory$ and $\grid$) can outperform $\tile$, $\direct$, $\aodha$, and $\rbf$ on both POI type classification task and geo-aware image classification task. \citet{mai2020se} showed that $\theory$ can outperform $\direct$ on geographic question answering task.

	\item Distance Prevervation: In terms of the question whether a location encoder is distance preserved (Property \ref{prop:dist}), we mainly consider them from a empirical perspective, e.g., whether the response map of a location encoder shows a spatial continuity pattern or a spatial heterogeneity pattern. The former implies that this model is distance preserved, while the latter indicates otherwise. For example, \citet{mai2020multi} systematically compared the response maps of different location encoders such as $\tile$, $\direct$, $\aodha$, $\theory$, $\grid$, and $\rbf$, after training on the POI type classification task (See Figure 2 in \citet{mai2020multi}). The  results show that $\direct$, $\aodha$, $\theory$, and $\rbf$ are distance preserved while $tile$ and $\direct$ are not. Moreover, as we discussed in Section \ref{sec:multiscale}, $\theory$ also have a theoretical proof for distance preservation. So we put ``Yes$^+$'' in Table \ref{tab:review}.	
	In terms of other location encoders, since no experiment or theoretical proof has been done, whether they have this property is unknown.
	
	\item Direction Awareness: A similar logic is used here. The response maps produced by \citet{mai2020multi} showed that $\direct$ and $\theory$ is aware of direction information while $\rbf$ is not. No conclusion can be drawn for other models.
	
	\item \sagatmodel~ is a bit different. Its properties also depend on the property of the used $\enc(\cdot)$ in Equation \ref{equ:sagat_6}. \sagatmodel~ can be a parametric or non-parametric (e.g., use $\rbf$ as $\enc(\cdot)$ ) model. It becomes a multi-scale approach if $\enc(\cdot)$ uses a multi-scale representation. Whether \sagatmodel~ has the distance preservation and direction awareness property also depends on the used $\enc(\cdot)$. 
	
\end{enumerate}

\section{Applying location encoding to different types of spatial data} \label{sec:complex-geo}

Location encoders can be directly utilized on multiple point set-based GeoAI tasks such as geo-aware image classification, POI type classification, and point cloud segmentation. However, there are many other tasks that are defined on other types of spatial data such as polylines, polygons, and graphs (networks). This section discusses the potential of location encoders to model these types of spatial data.

\subsection{Polyline}   \label{sec:line}

The location-to-polyline relation can be seen as an analogy of the word-to-sentence relation.
In NLP, a sentence, as an ordered sequence of words, can be encoded by different sequential neural nets such as different recurrent neural networks (RNN) \citep{hochreiter1997long,cho2014learning}  
and Transformer \citep{vaswani2017attention}. 
Their idea is to feed the embedding of each word token into a sequential model at each time step to encode the whole word sequence as one single hidden state or a sequence of hidden states.

Similarly, we can encode a polyline as an ordered sequence of locations, by using these sequential neural network models. At each time step, we will encode the current location into a location embedding and feed it into the sequential model. In fact, several recent work about human mobility directly follow this idea. \citet{xu2018encoding} utilized a $\direct$ location encoder to represent each trajectory point into a location embedding. Then a trajectory, represented as a sequence of location embeddings, are encoded by an LSTM for pedestrian trajectory prediction. 
Similarly, \citet{rao2020lstm} proposed an LSTM-TrajGAN framework to generate privacy-preserving synthetic trajectory data in which each trajectory point was encoded by a $\direct$ location encoder.

\subsection{Polygon}   \label{sec:polygon}
Encoding polygon geometries into the 
embedding space is a logical next step. 
It is very useful for several geospatial tasks which require comparing polygon geometries such as geographic entity alignment \citep{trisedya2019entity}, spatial topological reasoning \citep{regalia2019computing}, and geographic question answering \citep{mai2019relaxing,mai2020se,mai2021geographic}.

However,  unlike 
a polyline, which can be represented by an ordered sequence of locations, a polygon 
should be represented by all locations within it. The topological relationships between any location $\bx$ and a polygon should be preserved after the polygon encoding process. In other words, a polygon encoder should be topology aware. 
As far as we know, 
polygon encoding is still an ongoing research problem that does not have satisfactory solutions. 
\citet{mai2020se} presented a geographic entity bounding box encoding model as the first step towards polygon encoding 
by uniformly sampling a location from within the bounding box of a geographic entity and feeding it to a location encoder. 
Despite its innovativeness, this model still cannot handle fine-grained polygon geometries. \citet{yan2021graph} proposed a graph convolutional autoencoder (GCAE) which can encode simple polygons, i.e., polygons that do not intersect themselves and have no holes. GCAE converts the exterior of a simple polygon into a graph and then encodes this graph into an embedding space. 
The shortcoming of GCAE is that it  
cannot handle polygons with holes and multipolygons. Moreover, it cannot preserve the topology information.
So one interesting future research direction is developing a topology-aware polygon encoder which can handle both simple and complex polygons.

\subsection{Graph}   \label{sec:graph}

Graph (or network) is also an important spatial data format 
used in multiple geospatial data sets such as transportation networks \citep{li2017diffusion,cai2020traffic}, spatial social networks \citep{andris2016integrating}, and geographic knowledge graph (GeoKG) \citep{mai2020se}. 
A graph can be defined as $\graph = (\nodeset, \edgeset)$ where $\nodeset$ and $\edgeset$ are the set of nodes and edges in this graph. 
In the geospatial domain, each node $\node \in \nodeset$ or a subset of nodes in $\nodeset$ is associated with a location $\bx$ \footnote{Each node can associate with more complex geometries, and one single location is a simplification.} such as the sensor locations in a sensor network, users' locations in a spatial social network, or 
locations of geographic entities in a GeoKG. We further call this 
kind of graph a \textit{spatially embedded graph}.

The early practice to encode 
spatially embedded graphs is to treat them as normal non-spatial graphs and use some existing GNN models 
or (knowledge) graph embedding models \citep{grover2016node2vec,bordes2013translating,trouillon2017knowledge}. 
In order to add the spatial information as additional features without significantly modifying the existing 
architectures, we can modify the node 
encoder by using one $\enc(\bx)$ in Table \ref{tab:review} as the node encoder or one component of it while keeping other components unchanged. 
The model can be trained with the same loss function.
\citet{mai2020se} adopted exactly this practice and developed a spatially-explicit knowledge graph embedding model.
Similar ideas can be applied to 
other spatially embedded graphs.

Interestingly, other than the normal graph data, many pioneer research applied GNN models to a point set through a \textit{point-set-to-graph} conversion. 
They first converted point set $\mP$ into a graph based on spatial relations, e.g., a k-th nearest neighbor spatial graph, in which nodes indicate points while edges are associated with pairwise distance based weights. After this conversion, a GNN model is applied on this graph so that node attribute prediction can be done based on not only the nodes' own features but also their spatial context. 
Many GeoAI research has adopted this practice to tackle different tasks
including air quality forecasting \citep{lin2018exploiting}, 
place characteristics prediction \citep{zhu2020understanding}, GeoNames entity embedding learning \citep{kejriwal2017neural}, and different spatial interpolation problems \citep{apple2020kriging,wu2020kriging}. 
We argue that \textit{this kind of distance weighted graph method is insufficient to capture the relative spatial relations}, since it necessarily forfeits information about the spatial layout of points. Some important spatial information is lost such as the direction relations which are important for certain tasks when isotropic assumption is not held. 
Instead, we advocate the idea of using any $\enc^{(\mP)}(\bx)$ discussed in Section \ref{sec:agg} for these tasks since $\enc^{(\mP)}(\bx)$ is better at capturing spatial relations among locations.

\subsection{Raster} \label{sec:raster}

Convolutional Nerual Networks (CNNs) \citep{lecun95convnet} are at the core of many highly successful models in manipulating raster data such as image classification, image generation, and image understanding. This great success is due to the ability of the convolution operation to exploit the principles of locality, stationarity, and compositionality. 
Locality is due to the local connectivity, stationarity is owed to shift-invariance, and compositionality stems from the multi-resolution structure of the raster data \citep{bronstein2017geometric}. The number of learnable parameters is greatly reduced because of its feature locality and weight sharing across the data domain \citep{valsesia2018learning}.
Due to the success of location encoding on vector data, it is particularly interesting to think about the questions \textit{how we can apply location encoding techniques on rasters} and \textit{what the  benefits are}.

Interestingly, with increasing popularity of the Transformer \citep{vaswani2017attention} architecture, 
several efforts have been made to replace CNN with a Transformer-like architecture for raster-based tasks. The idea is that instead of using CNN kernels, we first encode the pixel features as well as pixel locations into the embedding space with a location-encoder-like architecture, so-called \textit{pixel position encoding}, and then a self-attention is applied on top of these pixel embeddings for different vision tasks.
However, one problem with this approach is that this per-pixel based self-attention has a very high computational cost.
One solution, among others, proposed by Vision Transformer (ViT) \citep{dosovitskiy2021image} 
uses a per-image-patch (instead of per-pixel ) self-attention which significantly lowers the computational cost. \citet{dosovitskiy2021image} showed that ViT can outperform traditional CNN-based models on several image classification benchmarks. 
However, the patch size becomes an important hyperparameter which will significantly affect model performance. 
Applying location encoders on raster data is a very new research direction. Existing work mainly focuses on encoding the positions of pixels on an image. When a pixel represents an area on the earth's surface (e.g., pixels in a satellite image), 
it is potentially very beneficial to encode pixel's geographic locations rather than its image positions. The geo-locations can serve as a channel, which transfers knowledge learnt from large quantities of unlabeled data (geographic data, geo-tagged image, or geo-tagged text) to the supervised learning tasks.

\section{Conclusion and Vision for Future Work}
\label{sec:future}

In this work, we 
formulate location encoding as an inductive learning based, task agnostic encoding technique for geographic locations. 
A formal definition of location encoding is provided, and
two expected properties -- distance preservation and direction awareness -- are discussed from the perspective of GIScience.
We illustrate the necessity of location encoding for GeoAI from a statistical machine learning perspective. 
A general classification framework has been provided to understand the current landscape of location encoding research (See Table \ref{tab:review}). 
We classify the existing location encoders into two categories: single point location encoder $\enc(\bx)$ and aggregation location encoder $\enc^{(\mP)}(\bx)$. 
For each category, we unify the location encoders into the same formulation framework (See Equations \ref{equ:locenc} and \ref{equ:aggenc}). Different location encoders are also compared based on various characteristics.
Finally, we demonstrate the possible usage of location encoding for different types of spatial data.

There are several interesting future research directions of location encoding:
\begin{enumerate}
	\item \textbf{Region representation learning}: As we
	discussed in Section \ref{sec:polygon}, there is no satisfactory solution for 
	polygon encoding (so called region representation learning), which will be very useful in various tasks such as geographic entity alignment and topological relation reasoning. 
	How to design a topology-aware polygon encoder which can handle simple polygons, polygon with holes, and multipolygons simultaneously is an interesting future research direction.

	\item \textbf{Spatiotemporal point encoding}: All the methods discussed so far are focused on location information whereas the temporal aspect of geospatial data is also very important. 
		Several important related questions are: 
		1) How to utilize temporal information in GeoAI models? 
		2) Can we encode temporal information in a similar manner as spatial information? 
		3) What are the important properties we need to preserve when doing temporal encoding? 
		4) How to combine temporal encoding and location encoding in a single framework?
		As for event sequences that happen synchronously \citep{kazemi2019time2vec}, i.e., sampled at regular intervals, the temporal information can be modeled implicitly by RNNs, or fed in RNNs as another input dimension after transforming time into handcrafted features \citep{du2016recurrent,li2017time,rao2020lstm}.
		Instead of using handcrafted temporal features, recent work proposed to encode time as learnable 
		vector representations such as Time2Vec \citep{kazemi2019time2vec} and \citet{cai2020traffic}. These temporal encoders are expected to preserve important properties such as periodicity, temporal continuity, invariance to time rescaling, and so on. 
		However, there are no systematic comparison studies among these temporal encoding approaches.
		As for combining location and temporal encoding, one obvious way is to add temporal information as an additional dimension of the location features. 
		\citet{mac2019presence} adopted this practice by adding time as an additional feature of $\psenc_{\aodha}(\bx)$ in Equation \ref{equ:wrap}.
		This leaded to a small performance improvement (0.25\%-1.37\%).
		However, they failed to consider those important properties of time mentioned above.
		Future research is needed to study the pros and cons of different temporal encoding approaches and how to combine it with location encoding.

	\item \textbf{Spherical location encoding}: As we discussed in Section \ref{sec:raster}, currently, there are no existing location encoders which can preserve spherical surface distance.
	When we are dealing with large-scale geospatial data sets (e.g., global SST data, species occurrences all over the world) 
	in which the map distortion problem is no longer negligible, 
	a spherical-aware location encoder is required which enable us to directly \textit{calculate on a round planet} \citep{chrisman2017calculating}.

	\item \textbf{Unsupervised learning for location encoding}: Most of the location encoders listed in Table \ref{tab:review} are trained in a supervised learning fashion which prohibits the application of the trained location embedding on other tasks. In contrast, text encoding methods, e.g., BERT, 
	are trained in an unsupervised 
	manner from numerous unlabeled data, and the pretrained model can be utilized in different downstream tasks \citep{devlin2018bert}. 
	How to design an unsupervised learning framework for location encoding 
	is a very attractive research direction. Recently, multiple point cloud generative models have been proposed such as 
	r-GAN/l-GAN \citep{achlioptas2018learning},  
	Graph-Conv GAN \citep{valsesia2018learning}, tree-GAN \citep{shu20193d}, 
	PointFlow \citep{yang2019pointflow}, and Generative PointNet \citep{xie2021generative}. 
	Their objective is to reconstruct given point clouds. 
	This presents one possible unsupervised learning framework of location encoding for unmarked points (points without attributes).
	Another interesting idea is unsupervised learning of the spatial distribution of marked points (points with attributes).
	
\end{enumerate}

\textbf{Data and Codes Availability Statement}

There is no code implementation for this review paper.

\section*{Acknowledgements}

We would like to thank Dr. Fei Du for his comments on the differences between location encoding and geohash. We also want to Thank Prof. Stefano Ermon for his suggestions on unsupervised learning for location encoding.
We would like to thank the three anonymous reviewers for their thoughtful comments and suggestions. 

This work is mainly funded by the National Science Foundation under Grant No. 2033521 A1 -- KnowWhereGraph: Enriching and Linking Cross-Domain Knowledge Graphs using Spatially-Explicit AI Technologies. Gengchen Mai acknowledges the support by UCSB Schmidt Summer Research Accelerator Award, Microsoft AI for Earth Grant, and the Office of the Director of National Intelligence (ODNI), Intelligence Advanced Research Projects Activity (IARPA), via 2021-2011000004. 
Yingjie Hu acknowledges support by the National Science Foundation under Grant No. BCS-2117771. Any views, opinions, findings, and conclusions or recommendations expressed in this material are those of the authors and should not be interpreted as necessarily representing the official policies, either expressed or implied, of NSF, ODNI, IARPA, or the U.S. Government. The U.S. Government is authorized to reproduce and distribute reprints for governmental purposes not-withstanding any copyright annotation therein.

\section*{Notes on contributors}

\textbf{Gengchen Mai} is a Postdoctoral Scholar at the Stanford AI Lab, Department of Computer Science, Stanford University. He is also affiliated with the Stanford Sustainability and AI Lab. He obtained his Ph.D. degree in Geography from the Department of Geography, University of California, Santa Barbara in 2021 and B.S. degree in GIS from Wuhan University in 2015. His research mainly focuses on Spatially-Explicit Machine Learning, Geospatial Knowledge Graph, and Geographic Question Answering. The first draft of this manuscript was submitted when Gengchen was a Ph.D. student at the Space and Time for Knowledge Organization Lab, Department of Geography, UC Santa Barbara.

\textbf{Krzysztof Janowicz} is a Professor for Geoinformatics at the University of California, Santa Barbara and director of the Center for Spatial Studies.  His research focuses on how humans conceptualize the space around them based on their behavior, focusing particularly on regional and cultural differences with the ultimate goal of assisting machines to better understand the information needs of an increasingly diverse user base. Janowicz's expertise is in knowledge representation and reasoning as they apply to spatial and geographic data, e.g., in the form of knowledge graphs.

\textbf{Yingjie Hu} is an Assistant Professor in GIScience at the Department of Geography, University at Buffalo, The State University of New York. He holds a Ph.D. in Geography from the University of California, Santa Barbara. His main research interests include Geospatial Artificial Intelligence and Spatial Data Science.

\textbf{Song Gao} is an Assistant Professor in GIScience at the Department of Geography, University of Wisconsin-Madison. He holds a Ph.D. in Geography at the University of California, Santa Barbara. His main research interests include Place-Based GIS, Geospatial Data Science, and GeoAI approaches to Human Mobility and Social Sensing.

\textbf{Bo Yan} received his bachelor’s degree in GIS from Wuhan University and his PhD in Geography from the University of California Santa Barbara. Dr Yan’s research area includes Geospatial Knowledge Graphs and machine learning in the geospatial domain.

\textbf{Rui Zhu} is a Postdoctoral Scholar at the Center for Spatial Studies, University of California, Santa Barbara. He obtained his Ph.D. in Geography from the same university in 2020; his M.S. degree was in Information Science from the University of Pittsburgh. Zhu's research interests include geospatial semantics, spatial statistics, as well as their broader interactions in geospatial artificial intelligence (GeoAI). 

\textbf{Ling Cai} is a fourth-year PhD student at the Space and Time for Knowledge Organization Lab, Department of Geography, University of California, Santa Barbara. She obtained her M.S. degree in Geographical Information Science from Chinese Academy of Sciences and B.S. degree from Wuhan University. Her research interests include qualitative spatial temporal reasoning, temporal knowledge graph, neuro-symbolic AI, and urban computing.

\textbf{Ni Lao} is currently a research scientist at Google. He holds a Ph.D. degree in Computer Science from the Language Technologies Institute, School of Computer Science at Carnegie Mellon University. He was the Chief scientist and co-founder of Mosaix.ai, a voice search AI startup. He is an expert in Machine Learning (ML), Knowledge Graph (KG) and Natural Language Understanding (NLU). He is known for his work on large scale inference and learning on KGs. This paper was done when Ni worked at Mosaix.ai.

\bibliographystyle{tfv}


\end{document}